\documentclass[lettersize,journal]{IEEEtran}
\usepackage{amsmath,amsfonts}
\usepackage{algorithmic}
\usepackage{algorithm}
\usepackage{array}
\usepackage[caption=false,font=normalsize,labelfont=sf,textfont=sf]{subfig}
\usepackage{hyperref}
\hypersetup{%
  colorlinks=true,%
  linkcolor={red!50!black},
  citecolor={blue!65!black},
  urlcolor={blue!80!black}
  bookmarksnumbered=true,%
  bookmarksopen=true}
\usepackage{textcomp}
\usepackage{stfloats}
\usepackage{url}
\usepackage{verbatim}
\usepackage{graphicx}         
\usepackage{cite}
\usepackage{pifont}
\usepackage[table,xcdraw]{xcolor}  
\usepackage[percent]{overpic}     
\usepackage{colortbl}             
\usepackage{booktabs}             
\usepackage{caption}             
\usepackage{multirow}
\definecolor{customPurple}{RGB}{214, 216, 235}
\definecolor{FIRST}{RGB}{175, 0, 0}
\definecolor{SECOND}{RGB}{17, 17, 174}
\definecolor{THIRD}{RGB}{55, 127, 153}
\definecolor{REVISION}{RGB}{69, 115, 222}
\begin{document}

\title{Towards Remote Sensing Change Detection with Neural Memory}

\author{
Zhenyu Yang,
Gensheng Pei,
Yazhou Yao,
Tianfei Zhou,
Lizhong Ding,
Fumin Shen

\thanks{Z. Yang, G. Pei, and Y. Yao are with the School of Computer Science and Engineering, Nanjing University of Science and Technology, China.}
\thanks{T. Zhou and L. Ding are with the School of Computer Science and Technology, Beijing Institute of Technology, China.}
\thanks{F. Shen is with the School of Computer Science and Engineering, University of Electronic Science and Technology of China, China.}
\thanks{Corresponding author: Yazhou Yao.}
}

\markboth{}%
{Shell \MakeLowercase{\textit{et al.}}: A Sample Article Using IEEEtran.cls for IEEE Journals}


\maketitle

\begin{abstract}
Remote sensing change detection is essential for environmental monitoring, urban planning, and related applications. However, current methods often struggle to capture long-range dependencies while maintaining computational efficiency. Although Transformers can effectively model global context, their quadratic complexity poses scalability challenges, and existing linear attention approaches frequently fail to capture intricate spatiotemporal relationships. Drawing inspiration from the recent success of Titans in language tasks, we present \textbf{ChangeTitans}, the Titans-based framework for remote sensing change detection.
Specifically, we propose \textbf{VTitans}, the first Titans-based vision backbone that integrates \textbf{neural memory} with segmented local attention, thereby capturing long-range dependencies while mitigating computational overhead. Next, we present a \textbf{hierarchical VTitans-Adapter} to refine multi-scale features across different network layers. Finally, we introduce \textbf{TS-CBAM}, a two-stream fusion module leveraging cross-temporal attention to suppress pseudo-changes and enhance detection accuracy.
Experimental evaluations on four benchmark datasets (LEVIR-CD, WHU-CD, LEVIR-CD+, and SYSU-CD) demonstrate that ChangeTitans achieves state-of-the-art results, attaining \textbf{84.36\%} IoU and \textbf{91.52\%} F1-score on LEVIR-CD, while remaining computationally competitive.
Our code and model are available at {https://github.com/ChangeTitans/ChangeTitans}.
\end{abstract}


\section{Introduction}
\IEEEPARstart{R}{emote} sensing change detection (RSCD) is fundamental to environmental conservation, urban development, and disaster assessment \cite{zheng2023changen,bernhard2023mapformer,mall2023change,yao2023automated}. By contrasting multi‑temporal satellite or aerial images, RSCD seeks to highlight meaningful land‑surface changes while suppressing nuisance variations introduced by illumination, phenology, or sensor discrepancies. Achieving this goal in high‑resolution imagery demands models that are simultaneously context‑aware and computationally efficient, a balance that remains elusive despite recent progress in deep learning.

\begin{figure}[t]
\centering
\includegraphics[width=1.0\linewidth]{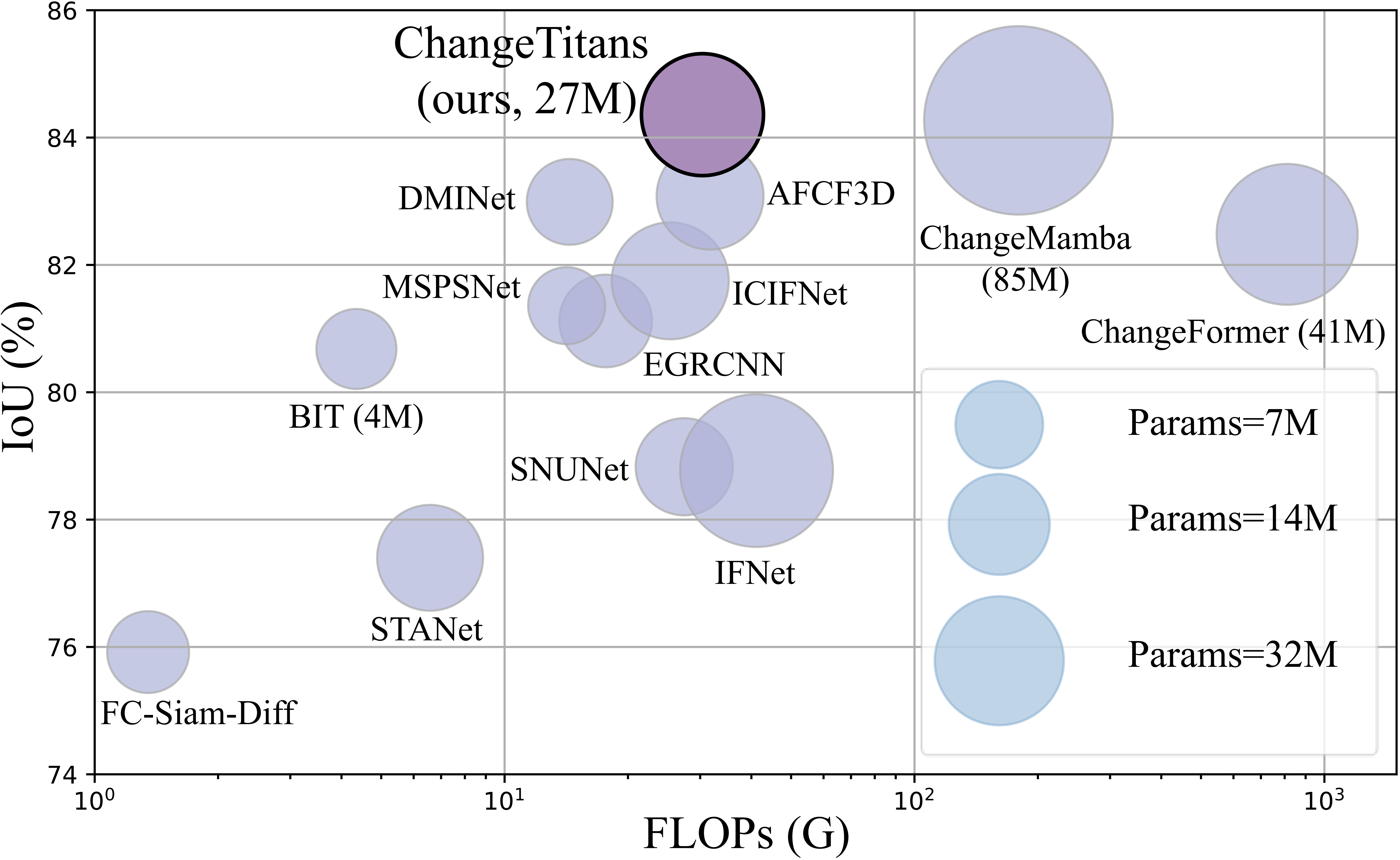}
\caption{Comparison of IoU on the LEVIR-CD dataset against computational efficiency (FLOPs) and model size (Parameters). The proposed ChangeTitans (\textcolor[rgb]{0.655, 0.553, 0.722}{\ding{108}}) achieves 84.36\% IoU with only 30.39G FLOPs, surpassing existing methods in both detection accuracy and computational efficiency.}
\label{fig:blob}
\end{figure}

Early convolutional neural‑network (CNN) approaches \cite{cai2024poly,ye2023adjacent,cai2025cycle,11311163,yao2021jo} extract local patterns effectively, yet their inherently limited receptive fields hinder long‑range reasoning across large scenes. Transformer‑based methods \cite{chen2021remote,bandara2022transformer,zhang2022swinsunet,ZHANG2024102425,yao2021non} overcome this limitation by employing self‑attention to model global relationships, achieving state‑of‑the‑art accuracy. Unfortunately, their quadratic complexity with respect to image resolution renders them impractical for widespread RSCD deployment. Linearized attention variants \cite{mamba,mamba2,10105896} alleviate the computational burden, but often at the expense of faithfully capturing intricate spatiotemporal dependencies.
A representative example is ChangeMamba \cite{chen2024changemamba}, which embeds state‑space models within a linear‑complexity framework. While effective, its reliance on sparse self‑attention and low‑rank approximations limits its ability to model very long contexts, prompting the addition of a heavy spatio‑temporal relationship module. As shown in Fig.~\ref{fig:blob}, the model ends up with 84.70M parameters and 179.32G FLOPs, far beyond typical resource limits, failing to meet the original goal of a lightweight design.

Titans \cite{behrouz2024titans} introduces a novel framework that bridges the gap between short-range precision and long-range memory in sequence modeling. Though highly effective for modeling local relationships, traditional attention mechanisms struggle with scalability due to their limited context window. In contrast, neural memory offers persistent long-range information retention but often at the cost of local detail. Titans provides a fresh perspective. By coupling neural memory with segmented attention, Titans retains historical information over arbitrarily long sequences while applying accurate local attention within each segment. 
Titans unifies these strengths, achieving fidelity and efficiency in language modeling, a property we seek to harness for vision.

Neural memory captures long-range context, while segmented attention focuses on local structure. This allows efficient reasoning over high-resolution scenes without quadratic complexity.
Guided by this principle, we present ChangeTitans, the first Titans‑style vision architecture for remote sensing change detection. Its backbone, VTitans, follows a flat Vision‑Transformer layout: high‑resolution images are divided into segments, intra‑segment attention refines fine textures, and distilled segment tokens are written to an external memory so that global cues can be revisited in \(\mathcal{O}(Nd\log N)\) time. Despite its non‑hierarchical design, VTitans matches peer Transformers on ImageNet‑1K \cite{deng2009imagenet}, ensuring that the features passed downstream are both semantically rich and globally coherent.
Flat representations are too coarse for the subtle spectral and structural cues that differentiate land‑cover classes. A lightweight VTitans‑Adapter, inspired by the design of \cite{duan2024vision}, converts the single‑scale VTitans output into a multi‑scale pyramid through shallow cross‑layer projections, injecting scale inductive bias while adding only marginal parameters. The resulting hierarchy simultaneously highlights crop‑scale details and landscape‑scale context, an essential property when changes may range from a single building footprint to an entire floodplain.
To compare bi‑temporal observations, we devise a Two‑Stream Convolutional Block Attention Module (TS‑CBAM). Earlier pipelines often rely on subtraction or fixed similarity metrics, which falter under seasonal shifts, illumination variation, or cross‑sensor noise \cite{dong2024changeclip}. TS‑CBAM aligns the two streams with coordinated channel‑wise and spatial attentions, letting the network learn where (spatial mask) and what (spectral–textural channels) to focus on. This joint modeling sharply reduces false positives while preserving genuine change boundaries, completing a cohesive pipeline in which memory, adapter, and attention each tackle a distinct RSCD challenge, yet operate in concert under practical computational budgets.

Our main contributions are as follows:
\begin{itemize}
    \item We present ChangeTitans, the first RSCD framework that natively transfers Titans-style neural memory and segmented attention to high‑resolution vision tasks, closing the gap between long‑range context modeling and practical efficiency.

    \item We propose VTitans, an efficient and memory‑augmented backbone that achieves competitive ImageNet‑1K accuracy while preserving linear complexity, furnishing globally consistent yet detail‑preserving features for downstream change analysis. 
    
    \item Our VTitans-Adapter builds a multi‑scale feature pyramid at negligible cost, and our TS‑CBAM leverages cross‑temporal attentions to disentangle true structural changes from spurious variations.
    
    \item Extensive experiments on the LEVIR-CD, WHU‑CD, LEVIR-CD+, and SYSU-CD datasets demonstrate that ChangeTitans reaches 84.36\% IoU and 91.52\% F1-score on LEVIR-CD with only 30.39G FLOPs, establishing a new state-of-the-art while markedly reducing computational cost.
\end{itemize}

\section{Related Works}

\textbf{Efficient Long-Range Modeling.}
The quadratic cost of vanilla self‑attention has motivated numerous efficiency‑oriented variants. Alternative paradigms depart from attention altogether: Mamba \cite{mamba,mamba2} adopts gated state‑space models (SSMs), and RWKV \cite{peng2023rwkv} merges recurrent updates with Transformer blocks.
In this work, our proposed ChangeTitans retains the expressive local modeling of attention but pushes scalability further by coupling segmented attention with a vision‑adapted neural memory. This yields linear complexity without stochastic approximations, recurrent bottlenecks, or heavy low‑rank projections, and it does so directly on high‑resolution imagery rather than one‑dimensional sequences.

\textbf{Remote Sensing Change Detection}.
Transformer‑based designs increasingly supplant CNNs in RSCD. BIT \cite{chen2021remote} pioneers a bitemporal Transformer encoder–decoder, ChangeFormer \cite{bandara2022transformer} introduces a hierarchical encoder with an MLP decoder, and MSCANet \cite{liu2022cnn} fuses CNN features with Transformer context aggregation. FHD \cite{pei2022feature} employs a MiT \cite{xie2021segformer} backbone and hierarchical differentiation, MapFormer \cite{bernhard2023mapformer} injects pre‑change semantics for feature refinement, and ChangeMamba \cite{chen2024changemamba} leverages state‑space encoders augmented by three spatio‑temporal modules.
In contrast, ChangeTitans delivers global context, multi‑scale awareness, and temporal alignment within a single memory‑augmented backbone plus two lightweight add‑ons (VTitans‑Adapter and TS‑CBAM), achieving state‑of‑the‑art accuracy with fewer parameters and one‑third of the FLOPs required by recent Transformer or SSM‑based counterparts.

\section{Method}

\subsection{Preliminaries}

Neural memory mechanisms in Titans \cite{behrouz2024titans}, inspired by RNNs \cite{Elman1990Finding}, use hidden states to retain historical context, enabling long-range modeling:
\begin{equation} \mathcal{M}_t=F(\mathcal{M}_{t-1};x_t). \end{equation}

The model compresses past inputs $x_1,\dots,x_{t-1}$ into a long-term memory module $\mathcal{M}_t$. To enhance memory efficiency, it employs a surprise metric, defined as the gradient of the loss w.r.t. the input:
\begin{equation} \nabla l(\mathcal{M}_{t-1};x_t). \end{equation}

A larger gradient indicates significant deviation from past data, prompting memory updates:
\begin{equation} \mathcal{M}_t=\mathcal{M}_{t-1}-\theta_t\nabla l(\mathcal{M}_{t-1};x_t), \end{equation}
where $\theta_t$ controls the update magnitude. However, large updates can overwrite useful historical information. To mitigate this, the surprise metric is decomposed into past surprise $S_{t-1}$ and momentary surprise:
\begin{equation} S_t=\eta_t S_{t-1}-\theta_t\nabla l(\mathcal{M}_{t-1};x_t),\quad \mathcal{M}_t=\mathcal{M}_{t-1}+S_t. \end{equation}

Here, $\eta_t$ serves as a decay factor, allowing adaptive retention of past surprises. When $\eta_t\rightarrow0$, past surprises are ignored; when $\eta_t\rightarrow1$, prior context is fully integrated.

To further refine memory management, a forgetting mechanism is introduced:
\begin{equation} \mathcal{M}_t=(1-\alpha_t)\mathcal{M}_{t-1}+S_t. \label{eq:5} \end{equation}
The forgetting rate $\alpha_t$ determines how much past memory is retained, where $\alpha_t\rightarrow0$ preserves historical information, and $\alpha_t\rightarrow1$ resets the memory module.

Memory updates are implemented using simple MLPs or other models. Given an input $x_t$, the memory loss is:
\begin{equation} l(\mathcal{M}_{t-1};x_t)=\left\|\mathcal{M}_{t-1}(k_t)-v_t\right\|_2^2, \label{eq:6} \end{equation}
where
\begin{equation} k_t=W_kx_t,\quad v_t=W_vx_t. \end{equation}

Here, $W_k,W_v\in\mathbb{R}^{d_\textrm{in}\times d_\textrm{in}}$ are linear transformations, treated as hyperparameters. The memory module $\mathcal{M}$ is updated in an inner loop like meta-learning \cite{nichol2018first, zintgraf2019fast}, while $W_k, W_v$ and other parameters are updated in the outer loop.

\begin{figure*}[t]
  \centering
  \includegraphics[width=\linewidth]{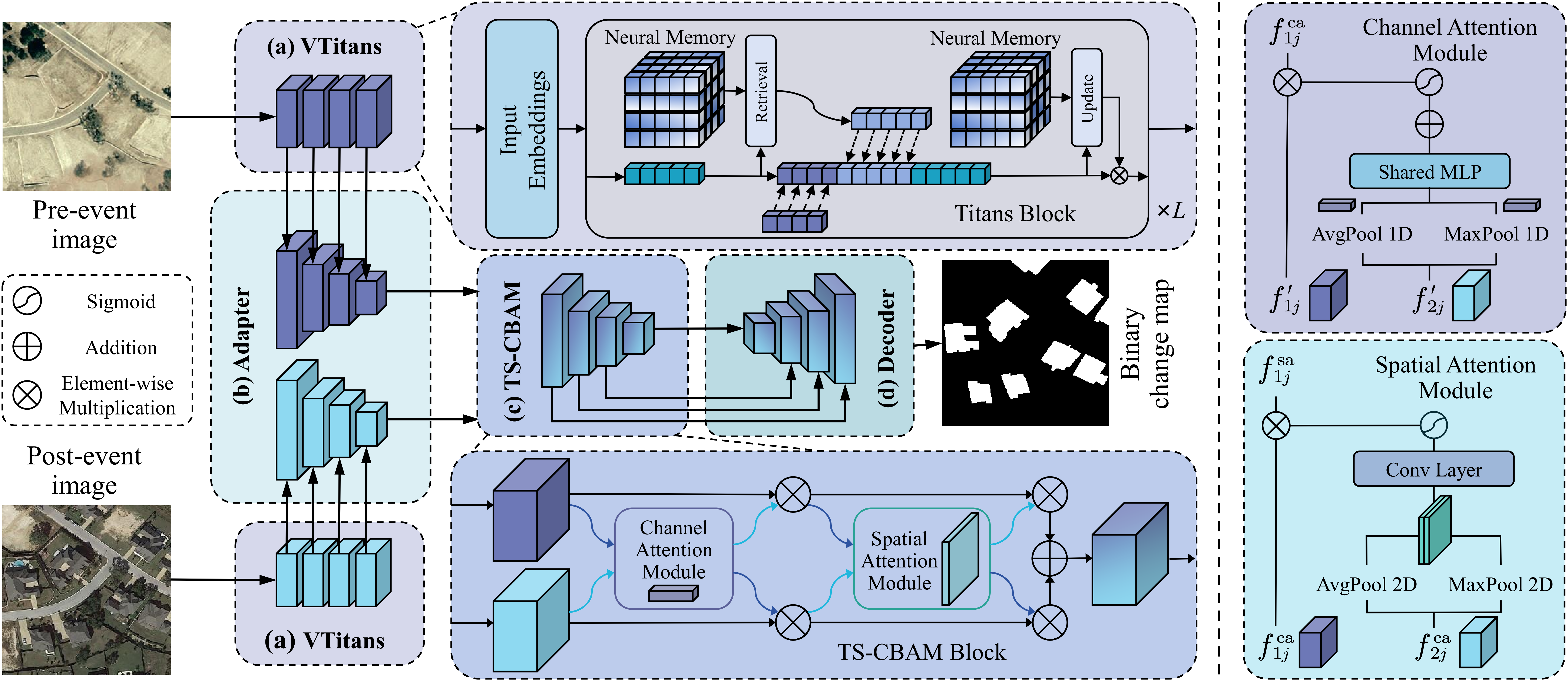}
  \caption{The overall architecture of the proposed ChangeTitans comprises four main components: \textbf{(a)} a Titans-based visual backbone (VTitans), \textbf{(b)} a lightweight VTitans-Adapter for constructing hierarchical feature representations, \textbf{(c)} a bi-temporal fusion module (TS-CBAM), and \textbf{(d)} a decoder for generating the final binary change maps. The internal structures of VTitans and TS-CBAM are illustrated in the upper and lower sub-blocks, respectively. Details of the VTitans-Adapter are provided in Fig.~\ref{fig:adapter}, while the Channel Attention and Spatial Attention modules within TS-CBAM are depicted on the right.}
  \label{fig:changetitans}
\end{figure*}

\subsection{Problem Statement}

In this paper, we address the change detection task in remote sensing. Specifically, we focus on binary change detection (BCD), where the goal is to generate a binary change map that identifies changed and unchanged regions. Formally, given a pair of co-registered remote sensing images $X_1,X_2\in\mathbb{R}^{H\times W\times C}$ acquired at different time points over the same geographical area, the objective is to learn a mapping function $\mathcal{F}:(X_1,X_2)\rightarrow Y$ where $Y\in\{0,1\}^{H\times W}$ is the binary change mask. Each pixel $(r,s)$ in $Y$ is assigned a value $Y_{r,s}=1$ if a change is detected at that location, and $Y_{r,s}=0$ otherwise. The primary challenge of BCD lies in accurately distinguishing meaningful changes, such as urban expansion or deforestation, from irrelevant variations caused by illumination differences, seasonal effects, or sensor noise. An ideal change detection model should be able to robustly extract discriminative features from $X_1$ and $X_2$, effectively suppress pseudo-changes, and generalize well across different datasets and geographical regions.

\subsection{Network Architecture}

The proposed ChangeTitans architecture is designed for efficient and accurate remote sensing change detection, leveraging neural memory and segmented attention mechanisms. As shown in Fig.~\ref{fig:changetitans}, the model consists of four key components: (a) \textbf{VTitans} backbone, a ViT-like \cite{dosovitskiy2020image} vision backbone based on the Titans architecture, integrating neural memory for long-range context modeling. (b) \textbf{VTitans-Adapter}, which transforms non-hierarchical VTitans features into multi-scale hierarchical representations, drawing inspiration from the works of \cite{duan2024vision}. (c) \textbf{TS-CBAM} fusion module, a two-stream attention mechanism for effectively fusing bi-temporal features while suppressing pseudo-changes, and (d) \textbf{Decoder}, a U-Net-inspired structure that reconstructs high-resolution change maps from fused features.

\begin{figure*}[t]
  \centering
  \includegraphics[width=\linewidth]{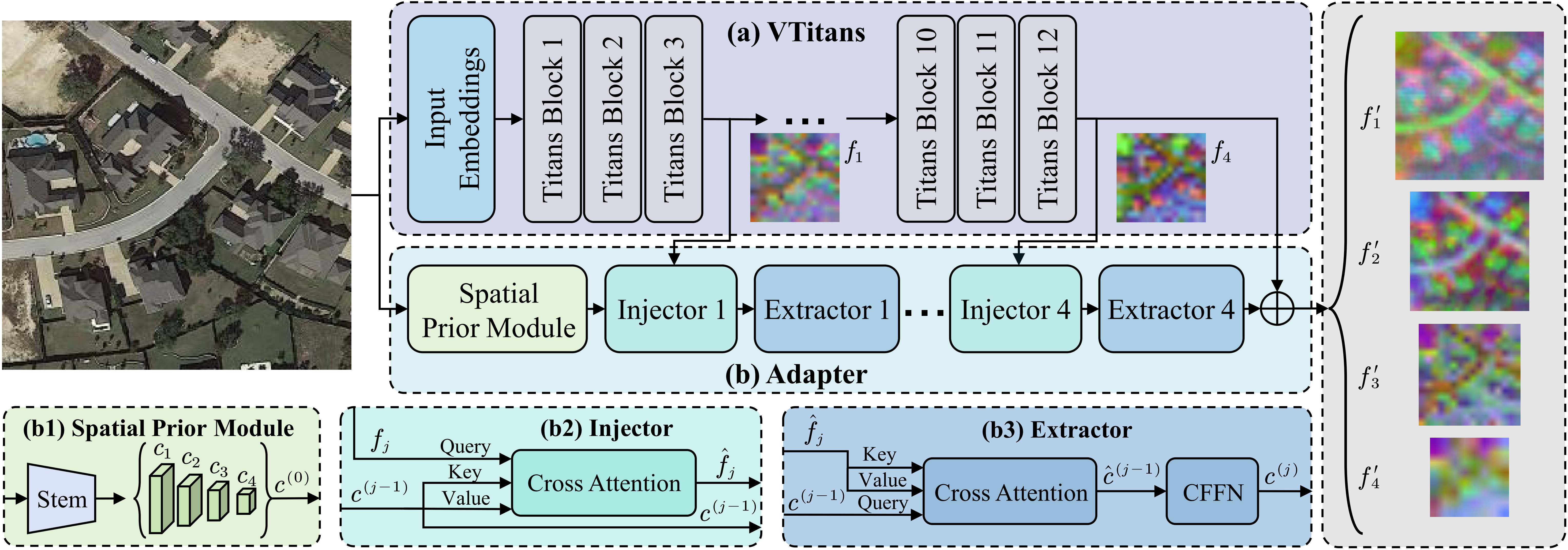}
  \caption{The structure of our VTitans-Adapter (\(L = 12\)) is illustrated as follows. It begins with \textbf{(b1)} the Spatial Prior Module to extract initial hierarchical features. These features are then progressively refined using four stages (\textit{i.e.}, Layers 3, 6, 9, and 12) of \textbf{(b2)} Injectors and \textbf{(b3)} Extractors, guided by the non-hierarchical representations from \textbf{(a)} VTitans. The final output serves as the encoder’s multi-scale representation. Feature maps are visualized by PCA.}
  \label{fig:adapter}
  \vspace{-0.1cm}
\end{figure*}

\begin{figure}[t]
  \centering
  \includegraphics[width=\linewidth]{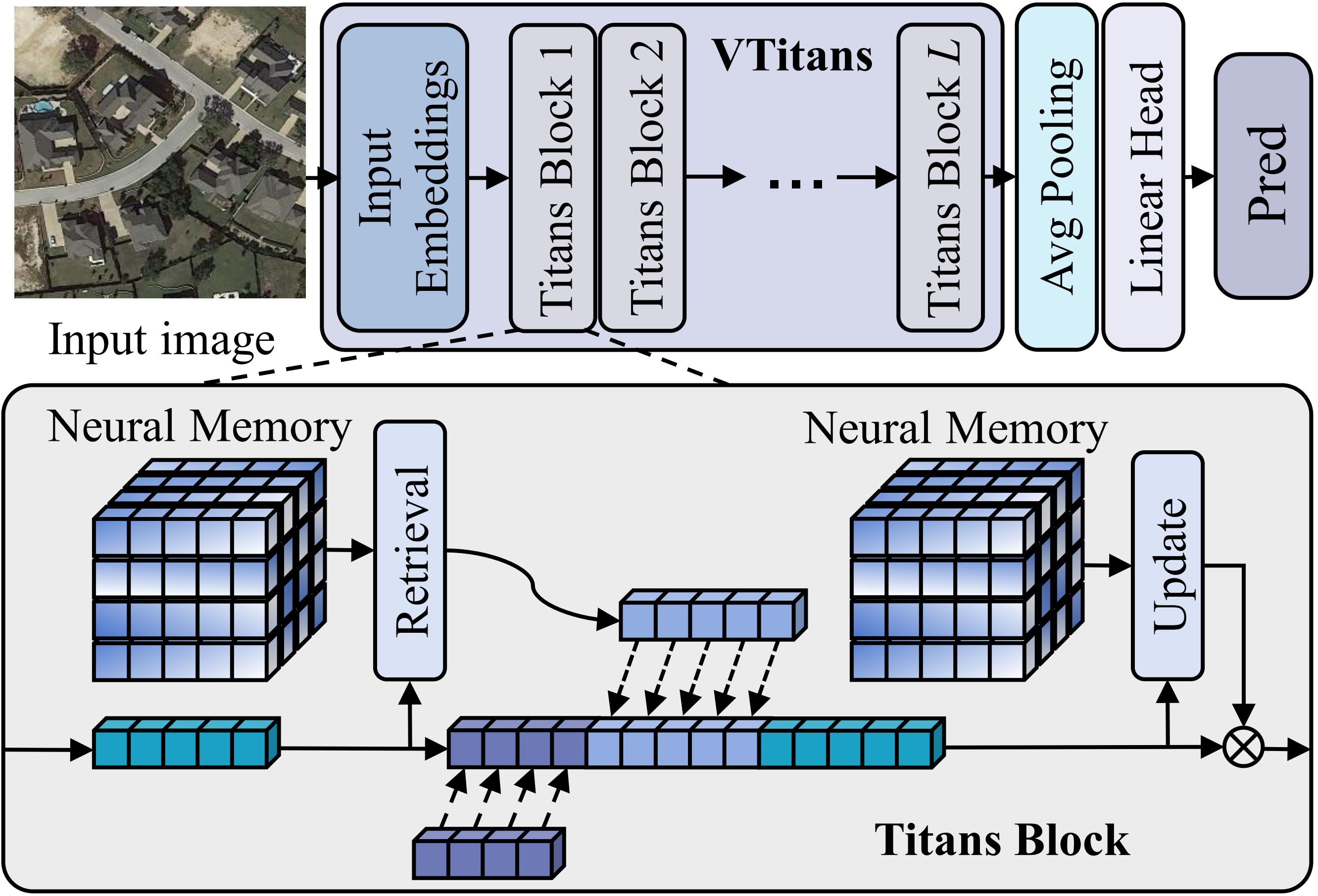}
  \caption{The overall architecture of the VTitans encoder. The input image is split into patches and passed through a series of Titans blocks, each equipped with segmented self-attention and neural memory. The memory modules enable long-range context modeling, while the segmented attention ensures computational efficiency. A linear prediction head follows the final block for downstream feature decoding.}
  \label{fig:vtitans}
  \vspace{-0.3cm}
\end{figure}

\textbf{Overview}. First, we transform the two-periods images $X_1,X_2\in\mathbb{R}^{H\times W\times C}$ into $T=HW/p^2$ patches, where $p$ denotes the patch size. Next, these patches undergo a linear mapping and are added with the positional encodings to obtain image tokens of shape $\mathbb{R}^{T\times C}$. These tokens are then fed into the VTitans encoder with $L$ layers, as shown in Fig.~\ref{fig:changetitans}(a). Since the feature maps extracted from VTitans are non-hierarchical, they may be limited in performance for dense prediction tasks such as change detection due to the lack of image prior information. Therefore, as illustrated in Fig.~\ref{fig:changetitans}(b), we extract feature maps $f_{i1},f_{i2},f_{i3},f_{i4}\in\mathbb{R}^{T\times C}$, $i\in\{1,2\}$ from every $L//4$ layers and transform them into hierarchical feature maps $f_{i1}^\prime\in\mathbb{R}^{(4H/p)\times(4W/p)\times C}$, $f_{i2}^\prime\in\mathbb{R}^{(2H/p)\times(2W/p)\times C}$, $f_{i3}^\prime\in\mathbb{R}^{(H/p)\times(W/p)\times C}$, and $f_{i4}^\prime\in\mathbb{R}^{(H/2p)\times(W/2p)\times C}$ using the parallel adapter structures. Next, we apply our proposed TS-CBAM to fuse the features from the two time periods, obtaining $f_j=\textrm{TS-CBAM}(f_{1j}^\prime,f_{2j}^\prime)$, where $j\in\{1,2,3,4\}$. Finally, the fused features are fed into the decoder to generate the binary change mask $\hat{Y}=\textrm{Decoder}(f_1,f_2,f_3,f_4)$. In the following subsections, we will provide a detailed description of each module.

\subsection{Titans-Based Vision Encoder}\label{sec:3.4}

We propose VTitans, a first Titans-based vision encoder for remote sensing change detection. As shown in Fig.~\ref{fig:changetitans}(a), VTitans consists of $L$ stacked Titans blocks. Given patch embeddings $x\in\mathbb{R}^{T\times C}$, each block processes fixed-size chunks $S^{(t)}$ and retrieves past information from memory:
\begin{equation} 
h_t=\mathcal{M}_{t-1}(q_t), \quad q_t=S^{(t)}W_q. 
\end{equation}

A learnable, input-independent persistent memory $P$ is then concatenated with the retrieved memory and input for self-attention:
\begin{equation} m_t=\textrm{SelfAttention}(P|h_t|S^{(t)}). \end{equation}

The updated memory is obtained via Eqs. \eqref{eq:5} and \eqref{eq:6}, and the final output is computed as:
\begin{equation} o_t=m_t\otimes \mathcal{M}_t(m_t), \end{equation}
where $\otimes$ denotes element-wise multiplication.

To transform non-hierarchical features into hierarchical ones, we introduce the VTitans-Adapter (see Fig.~\ref{fig:adapter}) in this work. First, the Spatial Prior Module (SPM) extracts multi-scale local features via convolutions. Next, the Spatial Feature Injector integrates spatial priors into VTitans via cross-attention, where VTitans features $f_j$ serve as queries, and spatial features $c^{(j-1)}$ act as keys/values:
\begin{equation} \hat{f}_j=f_j+\gamma_j^\textrm{in}\textrm{CrossAttention}(f_j,c^{(j-1)}). \end{equation}
For efficiency, deformable attention is used. The Multi-Scale Feature Extractor then updates spatial features through another cross-attention step:
\begin{equation} \hat{c}^{(j-1)}=c^{(j-1)}+\gamma_j^\textrm{ex}\textrm{CrossAttention}(c^{(j-1)},\hat{f}_j). \end{equation}
A Convolutional Feed-Forward Network (CFFN) enhances local continuity using depthwise convolutions:
\begin{equation} c^{(j)}=\hat{c}^{(j-1)}+\textrm{CFFN}(\hat{c}^{(j-1)}). \end{equation}

After four stages, spatial features incorporate VTitans information and are fused via interpolation for the final hierarchical representation:
\begin{equation} f_j^\prime=\textrm{Interpolate}(f_j)+c_j^\prime, \quad (c_1^\prime|c_2^\prime|c_3^\prime|c_4^\prime)=c^{(4)}. \end{equation}

This hierarchical representation effectively combines efficient attention from VTitans with spatial priors, making it well-suited for dense prediction tasks in remote sensing change detection.

\subsection{Two-stream Convolutional Block Attention}\label{sec:3.5}

Existing remote sensing change detection approaches often subtract features from two temporal images \cite{daudt2018fully} or apply fixed similarity metrics \cite{dong2024changeclip,zhoubocvpr}. These strategies can mislabel weather, lighting, or seasonal variations as genuine changes, leading to high false alarms. Additionally, minor misalignments in registration may generate spurious differences around boundaries or structures, further compromising accuracy.

To tackle these challenges, we propose a learnable TS-CBAM fusion module that leverages channel and spatial attentions across two temporal streams, thereby capturing deeper spatiotemporal patterns. As depicted in Fig.~\ref{fig:changetitans}(c), TS-CBAM draws upon the fundamental concept of gating mechanisms, originally seen in CBAM \cite{woo2018cbam}, but is re-architected to address the complexities of remote sensing change detection. By learning data-driven weight assignments and jointly modeling spatiotemporal cues, TS-CBAM effectively suppresses nuisances introduced by misregistration or environmental fluctuations and highlights truly informative change signals in the fused representation.

Building on this, our TS-CBAM enables cross-temporal feature interaction by using feature maps from one time period as the gating mechanism for the other, facilitating effective feature fusion. Given an input feature $f_{1j}^\prime,f_{2j}^\prime\in\mathbb{R}^{h\times w\times C}$, we first apply the Channel Attention Module (CAM), which assigns different weights across feature channels to emphasize important ones. Taking the feature stream of $f_{1j}^\prime$ as an example, we perform global max pooling and global average pooling along the spatial dimension of $f_{2j}^\prime$ to generate $f_{2j}^\textrm{avg},f_{2j}^\textrm{max}\in\mathbb{R}^C$. The two vectors are then passed through a shared fully connected layer to produce the channel attention weights:
\begin{equation}
M_\textrm{cam}(f_{2j}^\prime)=\sigma(W_1f_{2j}^\textrm{avg}+W_2f_{2j}^\textrm{max}),
\end{equation}
where \( W_1,W_2 \in \mathbb{R}^{C \times C} \), and \( \sigma \) denotes the sigmoid function. The resulting attention weights \( M_\textrm{cam}(f_{2j}^\prime) \) are applied to \( f_{1j}^\prime \) via channel-wise weighting to obtain the output of CAM:
\begin{equation}
f_{1j}^\textrm{cam}=M_\textrm{cam}(f_{2j}^\prime)\otimes f_{1j}^\prime.
\end{equation}

Next, the Spatial Attention Module (SAM) is applied to assign different weights to various spatial locations in the feature map, highlighting the most critical regions. Taking the feature stream of $f_{1j}^\textrm{cam}$ as an example, global max pooling and global average pooling are performed along the channel dimension on the auxiliary feature map $f_{2j}^\textrm{cam}$, generating $f_{2j}^\textrm{cam~avg},f_{2j}^\textrm{cam~max}\in\mathbb{R}^{h\times w}$. The two feature maps are then concatenated along the channel dimension and passed through a $3\times 3$ convolution to capture local spatial correlations, producing the spatial attention map:
\begin{equation}
M_\textrm{sam}(f_{2j}^\textrm{cam})=\sigma(\textrm{Conv}^{3\times3}(f_{2j}^\textrm{cam~avg}\|f_{2j}^\textrm{cam~max})).
\end{equation}

The generated attention weights $M_\textrm{sam}(f_{2j}^\textrm{cam})$ are then applied to the input feature map $f_{1j}^\textrm{cam}$ through position-wise weighting to obtain the output of SAM:
\begin{equation}
f_{1j}^\textrm{sam}=M_\textrm{sam}(f_{2j}^\textrm{cam})\otimes f_{1j}^\textrm{cam}.
\end{equation}

Finally, the outputs from the two time periods are summed to obtain the fused features of our proposed TS-CBAM:
\begin{equation}
f_j=f_{1j}^\textrm{sam}+f_{2j}^\textrm{sam}.
\end{equation}

\subsection{Binary Change Detection Decoder}\label{sec:3.6}

The proposed decoder progressively reconstructs the change map by restoring spatial resolution and refining the multi-scale features produced by VTitans. It incorporates upsampling operations combined with Titans blocks at each stage, ensuring that spatial details are recovered while preserving contextual information critical for remote sensing change detection. 

To further integrate encoder knowledge, skip connections couple the corresponding encoder outputs with the upsampled features, which are then passed through three consecutive Titans blocks for thorough refinement. This strategy avoids overreliance on global features alone and helps the network maintain continuity in object boundaries, a frequent challenge in RSCD.
A \(1\times1\) convolution is finally applied to reduce the channel dimension, followed by convex upsampling to restore the original image resolution. Convex upsampling has emerged as an effective technique in high-resolution vision tasks where spatial precision is crucial. Prior works have successfully applied convex combinations for feature upsampling in areas such as optical flow estimation \cite{pei2022eccv,10298026} and video frame interpolation \cite{pei2024cvpr,10105896}, where preserving boundary structure and fine detail is essential. These methods demonstrate that enforcing convexity constraints during upsampling can help suppress noise and maintain semantic consistency. 

However, despite its effectiveness, convex upsampling remains largely unexplored in the domain of remote sensing, particularly for change detection. This is surprising, given that change detection requires pixel-wise accuracy and sensitivity to small structural differences between bi-temporal images. Traditional upsampling approaches, such as bilinear interpolation or transposed convolutions, often result in blurred or misaligned change maps, especially along object boundaries or in heterogeneous landscapes. This motivates us to introduce Convex Upsampling into the pipeline.

Specifically, given a low-resolution (LR) feature map $\hat{y}\in\mathbb{R}^{(H/p)\times (W/p)\times C}$, our goal is to reconstruct a high-resolution (HR) output $\hat{Y}\in(0,1)^{H\times W}$ that aligns with the input image resolution. We achieve this via a learned convex interpolation mechanism, which is both structure-aware and data-adaptive.

\textbf{Convex weight prediction}. A shallow decoder head $\phi(\cdot)$ takes the feature map and predicts a set of spatially-varying interpolation weights:
\begin{equation}
    \tilde{W}=\phi(\hat{y}),\quad
    W_i(r,s)=\frac{\exp(\tilde{W}_i(r,s))}{\sum_{j=1}^{k^2}\exp(\tilde{W}_j(r,s))},
\end{equation}
ensuring the convexity constraint: $\sum_iW_i(r,s)=1$, $W_i(r,s)\geq0$. Here, $k$ is the size of the neighborhood used for interpolation, typically $k=3$.

\textbf{Feature reconstruction}. We extract a local $k\times k$ patch from the LR feature map around each mapped HR location using bilinear sampling:
\begin{equation}
    P(r,s)\in\mathbb{R}^{C\times k^2}.
\end{equation}

The upsampled feature at each HR location is then reconstructed as:
\begin{equation}
    \hat{Y}_{r,s}=\sum_{i=1}^{k^2}W_i(r,s)\cdot P_i(r,s).
\end{equation}

This convex combination ensures that the high-resolution representation remains within the semantic span of the LR features, while allowing adaptive weighting based on local context.

A sigmoid activation function then produces the final pixel-wise change predictions, culminating in a robust and spatially accurate remote sensing change detection pipeline.

\subsection{Loss Function}
To optimize the change detection model, a hybrid loss function combining Binary Cross Entropy (BCE) loss and Dice loss is employed to balance pixel-wise classification accuracy and spatial consistency. The BCE loss is defined as:
\begin{equation}
\mathcal{L}_\textrm{BCE}=-\frac{1}{N}\sum_{(r,s)}[Y_{r,s}\log (\hat{Y}_{r,s})+(1-Y_{r,s})\log (1-\hat{Y}_{r,s})],
\end{equation}
where $N$ is the number of pixels, $Y_{r,s}\in\{0,1\}$ is the ground truth label for pixel $(r,s)$, and $\hat{Y}_{r,s}\in(0,1)$ is the predicted probability. BCE loss penalizes misclassified pixels and encourages accurate pixel-wise predictions.

To address class imbalance and improve the spatial coherence of predicted change regions, the Dice loss is introduced as:
\begin{equation}
\mathcal{L}_\textrm{Dice}=1-\frac{2\sum_{(r,s)}Y_{r,s}\hat{Y}_{r,s}+\epsilon}{\sum_{(r,s)}Y_{r,s}+\sum_{(r,s)}\hat{Y}_{r,s}+\epsilon},
\end{equation}
where $\epsilon$ is a small constant added for numerical stability. Dice loss measures the overlap between the predicted and ground truth masks, encouraging the model to produce smooth and contiguous change regions.

The final loss function is defined as a weighted sum of BCE loss and Dice loss:
\begin{equation}
\mathcal{L}=\mathcal{L}_\textrm{BCE}+\lambda\mathcal{L}_\textrm{Dice},
\end{equation}
where $\lambda$ is hyperparameters that balance the contribution of each loss term. This combined loss function ensures both pixel-wise accuracy and structural consistency in the predicted change maps.

\section{Experiments}

\subsection{Experimental Setup}

\textbf{Datasets}: We evaluate the proposed method on four publicly available remote sensing change detection datasets: LEVIR-CD \cite{chen2020spatial}, WHU-CD \cite{hong2023multi}, LEVIR-CD+ \cite{shen2021s2looking}, SYSU-CD \cite{shi2021deeply} and SAR-CD \cite{alatalo2023improved}.
\textbf{LEVIR-CD} contains 637 very high-resolution (\(1024 \times 1024\) pixels, 0.5m/pixel) bitemporal image pairs collected from Google Earth, covering 20 regions in the United States between 2002 and 2018. It includes 31,333 annotated building change instances, making it a widely used benchmark for fine-grained urban change detection.  
\textbf{WHU-CD}, derived from the WHU Building dataset, consists of two aerial images of Christchurch, New Zealand (captured in 2012 and 2016 at 0.3m/pixel resolution), covering an area of 20.5 km$^2$. The dataset reflects building growth from 12,796 to 16,077 instances, emphasizing structural urban expansion.
\textbf{LEVIR-CD+} extends the LEVIR-CD dataset with 985 bitemporal image pairs acquired over 5 to 14 years. It captures both construction and demolition events, offering a broader temporal context for evaluating long-term building change detection. 
\textbf{SYSU-CD} is a large-scale, category-agnostic change detection dataset composed of 20,000 aerial image pairs (\(256 \times 256\) pixels, 0.5m/pixel) collected over Hong Kong from 2007 to 2014. It encompasses a wide range of change types, including urban construction, vegetation dynamics, and coastal reclamation, making it particularly suitable for general-purpose and heterogeneous change detection tasks.
\textbf{SAR-CD} is based on the validation partition of a dataset constructed using Sentinel-1 SAR imagery over North-East Finland. It contains 10,000 $512\times512$ bi-temporal SAR image pairs, which are cropped into $256\times256$ patches and re-divided into 32,000 training, 4,000 validation, and 4,000 test samples. All samples are labeled with pixel-wise change masks generated through simulated changes, enabling robust evaluation under SAR-specific challenges such as speckle noise.

\begin{table*}[t]
\caption{Comparison results on the LEVIR-CD \cite{chen2020spatial} and WHU-CD \cite{hong2023multi} datasets. All metrics are reported as percentages (\%). The top three results are marked in \textcolor{FIRST}{red} (1st), \textcolor{SECOND}{blue} (2nd), and \textcolor{THIRD}{green} (3rd), respectively.}
\label{tab:levir}
\setlength{\tabcolsep}{5.5pt}
\resizebox{\textwidth}{!}{%
\begin{tabular}{lccc|cccc|cccc}
\specialrule{1pt}{0pt}{0pt}
\rowcolor[HTML]{D6D8EB} 
\multicolumn{1}{c}{\cellcolor[HTML]{D6D8EB}} &
  \cellcolor[HTML]{D6D8EB} &
  \cellcolor[HTML]{D6D8EB} &
  \cellcolor[HTML]{D6D8EB} &
  \multicolumn{4}{c|}{\cellcolor[HTML]{D6D8EB}\textbf{LEVIR}} &
  \multicolumn{4}{c}{\cellcolor[HTML]{D6D8EB}\textbf{WHU-CD}} \\ \cline{5-12} 
\rowcolor[HTML]{D6D8EB} 
\multicolumn{1}{c}{\multirow{-2}{*}{\cellcolor[HTML]{D6D8EB}\textbf{Method}}} &
  \multirow{-2}{*}{\cellcolor[HTML]{D6D8EB}\textbf{Venue}} &
  \multirow{-2}{*}{\cellcolor[HTML]{D6D8EB}\textbf{Params (M)}} &
  \multirow{-2}{*}{\cellcolor[HTML]{D6D8EB}\textbf{FLOPs (G)}} &
  \textbf{IoU} &
  \textbf{F1} &
  \textbf{P} &
  \textbf{R} &
  \textbf{IoU} &
  \textbf{F1} &
  \textbf{P} &
  \textbf{R} \\ \hline
\rowcolor[HTML]{EFEFEF} 
FC-Siam-Diff \cite{daudt2018fully}          & ICIP 2018   & 4.38  & 1.35   & 75.92 & 86.31 & 89.53 & 83.31 & 41.66 & 58.81 & 47.33 & 77.66 \\
\rowcolor[HTML]{EFEFEF} 
FC-Siam-Conc \cite{daudt2018fully}          & ICIP 2018   & 4.99  & 1.55   & 71.96 & 83.69 & 91.99 & 76.77 & 49.95 & 66.63 & 60.88 & 73.58 \\
\rowcolor[HTML]{EFEFEF} 
SNUNet \cite{fang2021snunet}                & GRSL 2021   & 12.03 & 27.44  & 78.83 & 88.16 & 89.18 & 87.17 & 71.67 & 83.50 & 85.60 & 81.49 \\
\rowcolor[HTML]{EFEFEF} 
BIT \cite{chen2021remote}                   & TGRS 2021   & 3.55  & 4.35   & 80.68 & 89.31 & 89.24 & 89.37 & 72.39 & 83.98 & 86.64 & 81.48 \\
\rowcolor[HTML]{EFEFEF} 
EGRCNN \cite{bai2021edge}                   & TGRS 2022   & 9.64  & 17.64  & 81.12 & 89.59 & \textcolor{THIRD}{92.83} & 86.46 & 75.03 & 84.86 & 90.23 & 81.09 \\
\rowcolor[HTML]{EFEFEF} 
MSPSNet \cite{guo2021deep}                  & TGRS 2022   & 2.21  & 14.17  & 81.36 & 89.18 & 91.38 & 87.08 & 79.04 & 88.29 & 89.95 & 86.70 \\
\rowcolor[HTML]{EFEFEF} 
ICIFNet \cite{feng2022icif}                 & TGRS 2022   & 23.82 & 25.36  & 81.75 & 89.96 & 91.32 & 88.64 & 79.24 & 88.32 & 92.98 & 85.56 \\
\rowcolor[HTML]{EFEFEF} 
ChangeFormer \cite{bandara2022transformer}  & IGARSS 2022 & 41.02 & 811.15 & 82.48 & 90.40 & 92.05 & 88.80 & 71.91 & 83.66 & 85.49 & 81.90 \\
\rowcolor[HTML]{EFEFEF} 
TinyCD \cite{codegoni2023tinycd}            & NCA 2022    & 0.28  & 1.45   & 83.57 & 91.05 & 92.68 & 89.47 & 84.74 & 91.74 & 91.72 & 91.76 \\
\rowcolor[HTML]{EFEFEF} 
DMINet \cite{feng2023change}                & TGRS 2023   & 6.24  & 14.42  & 82.99 & 90.71 & 92.52 & 89.95 & 79.68 & 88.69 & 93.84 & 86.25 \\
\rowcolor[HTML]{EFEFEF} 
WNet \cite{tang2023wnet}                    & TGRS 2023   & 43.07 & 19.20  & 82.93 & 90.67 & 91.16 & 90.18 & 83.91 & 91.25 & 92.37 & 90.15 \\
\rowcolor[HTML]{EFEFEF} 
AFCF-3D \cite{ye2023adjacent}               & TGRS 2023   & 17.54 & 31.72  & 83.08 & 90.76 & 91.35 & 90.17 & 87.93 & 93.58 & 93.47 & \textcolor{SECOND}{93.69} \\
\rowcolor[HTML]{EFEFEF} 
VcT \cite{jiang2023vct}                     & TGRS 2023   & 3.57  & 10.64  & 81.89 & 90.04 & 92.57 & 87.65 & 81.12 & 89.58 & 89.39 & 89.77 \\
\rowcolor[HTML]{EFEFEF} 
CCLNet++ \cite{song2023toward}              & TGRS 2023   & 28.78 & 23.27  & 83.62 & 91.08 & 92.31 & 89.88 & 82.32 & 90.31 & 89.83 & 90.78 \\
\rowcolor[HTML]{EFEFEF} 
S$^2$CD \cite{wang2024summator}             & TGRS 2024   & 3.22  & 9.26   & 82.78 & 90.58 & 92.12 & 89.08 & 82.13 & 90.19 & 92.45 & 88.74 \\
\rowcolor[HTML]{EFEFEF} 
CF-GCN \cite{wang2024cf}                    & TGRS 2024   & 13.58 & 43.93  & 83.41 & 90.96 & 91.75 & 90.18 & 84.90 & 91.83 & \textcolor{SECOND}{94.81} & 89.04 \\
\rowcolor[HTML]{EFEFEF} 
SEIFNet \cite{huang2024spatiotemporal}      & TGRS 2024   & 8.37  & 27.91  & 83.40 & 90.95 & 92.49 & 89.46 & 76.04 & 86.39 & 87.01 & 85.77 \\
\rowcolor[HTML]{EFEFEF} 
ChangeCLIP \cite{dong2024changeclip}        & ISPRS 2024  & -     & -      & 83.99 & 91.30 & \textcolor{SECOND}{93.68} & 89.04 & \textcolor{FIRST}{90.08} & \textcolor{FIRST}{94.78} & \textcolor{FIRST}{96.02} & \textcolor{THIRD}{93.58} \\
\rowcolor[HTML]{EFEFEF} 
CBSASNet \cite{he2024cbsasnet}              & TGRS 2024   & 5.76  & 31.64  & \textcolor{THIRD}{84.14} & \textcolor{THIRD}{91.39} & 92.47 & \textcolor{SECOND}{90.33} & 86.08 & 92.52 & 93.93 & 91.15 \\
\rowcolor[HTML]{EFEFEF} 
ChangeMamba \cite{chen2024changemamba}      & TGRS 2024   & 84.70 & 179.32 & \textcolor{SECOND}{84.27} & \textcolor{SECOND}{91.37} & \textcolor{FIRST}{93.87} & 89.00 & \textcolor{THIRD}{88.02} & \textcolor{THIRD}{93.63} & 94.22 & 93.05 \\
\rowcolor[HTML]{EFEFEF} 
SFEARNet \cite{li2025sfearnet}              & TGRS 2025   & 5.56  & 4.65   & 83.23 & 90.85 & 91.43 & \textcolor{THIRD}{90.27} & 85.81 & 92.36 & 94.25 & 90.55 \\
\rowcolor[HTML]{EFEFEF} 
AMDANet \cite{su2025amdanet}                & TGRS 2025   & 25.83 & 77.23  & 82.34 & 90.32 & 92.45 & 88.28 & 80.68 & 89.30 & 88.77 & 89.84 \\
\hline
\rowcolor[HTML]{D6D8EB} 
\textbf{ChangeTitans (ours)}                & -           & 27.15 & 30.39  & \textcolor{FIRST}{84.36} & \textcolor{FIRST}{91.52} & 92.49 & \textcolor{FIRST}{90.56} & \textcolor{SECOND}{88.56} & \textcolor{SECOND}{94.10}  & \textcolor{THIRD}{94.33} & \textcolor{FIRST}{93.87} \\ 
\specialrule{1pt}{0pt}{0pt}
\end{tabular}}
\vspace{-0.1cm}
\end{table*}

\textbf{Implementation details}: We employ a sliding window approach to crop image pairs into $256\times256$ patches. The network is optimized using the Adam optimizer \cite{kingma2014adam} with an initial learning rate of 1e-5, a weight decay of 1e-4, and a batch size of 8 per GPU. The learning rate is scheduled using the cosine annealing strategy with a warm-up phase, where the warm-up period is set to 20 epochs with a multiplier of 10. Training is conducted for 200 epochs across all datasets, with gradient clipping set to 0.5. Following \cite{pei2022feature}, we apply data augmentation techniques, including random rotation, random flipping, random brightness contrast, and Gaussian blur.

\begin{table}[t]
\centering
\caption{Pre-training configuration for the VTitans encoder on ImageNet-1K \cite{deng2009imagenet}.}
\label{tab:conf}
\setlength{\tabcolsep}{15pt}
\resizebox{1.0\columnwidth}{!}{%
\begin{tabular}{lc}
\specialrule{1pt}{0pt}{0pt}
\rowcolor[HTML]{D6D8EB} 
\multicolumn{1}{c}{\cellcolor[HTML]{D6D8EB}\textbf{Setting}} & \textbf{Value} \\ \hline
\rowcolor[HTML]{EFEFEF} 
Number of Titans Blocks ($L$)\quad                           & 12             \\
\rowcolor[HTML]{EFEFEF} 
Embedding Dimension                                          & 192            \\
\rowcolor[HTML]{EFEFEF} 
Patch Size                                                   & 16×16          \\
\rowcolor[HTML]{EFEFEF} 
Chunk Size                                                   & 64             \\
\rowcolor[HTML]{EFEFEF} 
Memory Block Interval                                        & Every 3 layers \\
\rowcolor[HTML]{EFEFEF} 
Optimizer                                                    & Adam \cite{kingma2014adam}          \\
\rowcolor[HTML]{EFEFEF} 
Learning Rate                                                & 4e-3           \\
\rowcolor[HTML]{EFEFEF} 
Batch Size (per GPU)                                         & 64             \\
\rowcolor[HTML]{EFEFEF} 
Update Frequency                                             & 16             \\
\rowcolor[HTML]{EFEFEF} 
Input Resolution                                             & 224×224        \\
\rowcolor[HTML]{EFEFEF} 
Pre-training Dataset                                         & ImageNet-1K \cite{deng2009imagenet}    \\ \hline
\rowcolor[HTML]{D6D8EB} 
Top-1 Accuracy                                               & 75.70\%        \\ \specialrule{1pt}{0pt}{0pt}
\end{tabular}
}
\vspace{-0.2cm}
\end{table}

\begin{table}[t]
\caption{Comparison of vision backbone performance on ImageNet-1K \cite{deng2009imagenet} pre-training.}
\label{tab:pretrain}
\setlength{\tabcolsep}{3pt}
\resizebox{\columnwidth}{!}{%
\begin{tabular}{lcccc}
\specialrule{1pt}{0pt}{0pt}
\rowcolor[HTML]{D6D8EB} 
\multicolumn{1}{c}{\cellcolor[HTML]{D6D8EB}\textbf{Method}} &
  \textbf{Venue} &
  \textbf{\begin{tabular}[c]{@{}c@{}}Params\\ (M)\end{tabular}} &
  \textbf{\begin{tabular}[c]{@{}c@{}}FLOPs\\ (G)\end{tabular}} &
  \textbf{\begin{tabular}[c]{@{}c@{}}Top-1\\ Acc (\%)\end{tabular}} \\ \hline
\rowcolor[HTML]{ECF4FF} 
ResNet-18 \cite{he2016deep}                & CVPR 2016 & 11.7 & 1.8 & 69.9 \\
\rowcolor[HTML]{EFEFEF} 
MobileNetV2 x1.4 \cite{sandler2018mobilenetv2} & CVPR 2018 & 6.1 & 0.6 & 74.7 \\
\rowcolor[HTML]{EFEFEF} 
ShuffleNetV2 x2 \cite{ma2018shufflenet}    & ECCV 2018 & 7.4  & 0.6 & 74.9 \\
\rowcolor[HTML]{EFEFEF} 
ViT-Ti/16 \cite{dosovitskiy2020image}         & ICLR 2021 & 5.7  & 1.1 & 72.7 \\
\rowcolor[HTML]{EFEFEF} 
RepVGG-A1 \cite{ding2021repvgg}            & CVPR 2021 & 12.8 & 2.4 & 74.2 \\
\rowcolor[HTML]{EFEFEF} 
DeiT-T \cite{pmlr-v139-touvron21a}         & ICML 2021 & 5.7  & 1.3 & 72.2 \\
\rowcolor[HTML]{EFEFEF} 
ConViT-tiny \cite{d2021convit}             & ICML 2021 & 5.7  & 1.0 & 73.1 \\
\rowcolor[HTML]{EFEFEF} 
PiT-xs \cite{heo2021rethinking}            & ICCV 2021 & 10.6 & 1.4 & 72.4 \\
\rowcolor[HTML]{EFEFEF} 
CrossViT-Tiny \cite{heo2021rethinking}     & ICCV 2021 & 6.9  & 1.6 & 73.4 \\
\rowcolor[HTML]{EFEFEF} 
PVT-T \cite{wang2021pyramid}               & ICCV 2021 & 13.2 & 1.9 & 75.1 \\
\rowcolor[HTML]{EFEFEF} 
GFNet-Ti \cite{rao2021global}              & NeurIPS 2021 & 7.0 & 1.3 & 74.6 \\
\rowcolor[HTML]{EFEFEF} 
PoolFormer-S7 \cite{yu2022metaformer}      & CVPR 2022 & 8.6  & 1.1 & 73.0 \\
\rowcolor[HTML]{EFEFEF} 
DepViT-Lite-T \cite{ding2023visual}        & CVPR 2023 & 6.2  & 0.8 & 73.7 \\
\rowcolor[HTML]{EFEFEF} 
QuadMamba \cite{xie2024quadmamba}          & NeurIPS 2024 & 5.4 & 0.8 & 74.2 \\
\rowcolor[HTML]{EFEFEF} 
SHViT-S2 \cite{yun2024shvit}               & CVPR 2024 & 11.4 & 0.4 & 75.2 \\
\rowcolor[HTML]{EFEFEF} 
MobileMamba-T2 \cite{he2024mobilemamba}    & CVPR 2025 & 8.8  & 0.3 & 73.6 \\
\rowcolor[HTML]{EFEFEF} 
LSNet-T \cite{wang2025lsnet}               & CVPR 2025 & 11.4 & 0.3 & 74.9 \\
\rowcolor[HTML]{EFEFEF} 
EfficientViM-M2 \cite{lee2025efficientvim} & CVPR 2025 & 13.9 & 0.4 & 75.4 \\
\rowcolor[HTML]{EFEFEF} 
KAT-T \cite{yang2025kolmogorov}            & ICLR 2025 & 5.7  & 1.1 & 74.6 \\
\rowcolor[HTML]{EFEFEF} 
VRWKV-T \cite{duan2024vision}          & ICLR 2025 & 6.2  & 1.2 & 75.1 \\
\hline
\rowcolor[HTML]{D6D8EB} 
\textbf{VTitans (ours)}                    & -         & 11.7 & 1.9 & \textbf{75.7} \\ \specialrule{1pt}{0pt}{0pt}
\end{tabular}}
\vspace{-0.1cm}
\end{table}

\textbf{Evaluation Metrics}: To quantitatively assess the performance of RSCD models, we adopt four widely used metrics: Precision (P), Recall (R), F1-score (F1), and Intersection over Union (IoU). These metrics are defined as follows:

\begin{equation} \text{Precision} = \frac{TP}{TP + FP}, \end{equation} 
\begin{equation} \text{Recall} = \frac{TP}{TP + FN}, \end{equation} 
\begin{equation} \text{F1-score} = \frac{2 \times \text{Precision} \times \text{Recall}}{\text{Precision} + \text{Recall}}, \end{equation} 
\begin{equation} \text{IoU} = \frac{TP}{TP + FP + FN}, \end{equation}
where $TP$, $FP$, and $FN$ denote the numbers of true positives, false positives, and false negatives, respectively. All metrics are computed at the pixel level. These criteria provide a comprehensive evaluation of both detection accuracy and localization quality.
To further evaluate the boundary-preserving capability of the models, we additionally employ three boundary-sensitive metrics: \textit{Boundary F1 score} (BF1), which measures the alignment of predicted and ground-truth boundaries within a tolerance band; \textit{Trimap-based mIoU}, which focuses on prediction accuracy within narrow regions surrounding object boundaries; and \textit{Hausdorff distance} ($H$), which quantifies the maximum discrepancy between predicted and ground-truth contours. These metrics are defined as follows:
\begin{equation}
\text{BF1} = \frac{2 \times \text{Precision}_{b} \times \text{Recall}_{b}}{\text{Precision}_{b} + \text{Recall}_{b}},
\end{equation}
where $\text{Precision}_{b}$ and $\text{Recall}_{b}$ are computed by matching predicted and ground-truth boundaries within a tolerance band.  
\begin{equation}
\text{Trimap-mIoU} = \frac{TP_{t}}{TP_{t} + FP_{t} + FN_{t}},
\end{equation}
where $TP_{t}$, $FP_{t}$, and $FN_{t}$ denote true positives, false positives, and false negatives restricted to the trimap region around object boundaries.  
\begin{equation}
H(P, G) = \max \Big\{ \sup_{p \in P} \inf_{g \in G} d(p,g), \ \sup_{g \in G} \inf_{p \in P} d(g,p) \Big\},
\end{equation}
where $P$ and $G$ are the sets of predicted and ground-truth boundary pixels, respectively, and $d(\cdot,\cdot)$ denotes the Euclidean distance.  
Together, these metrics complement the standard pixel-level evaluations and provide a more holistic assessment of boundary quality in change detection.

\begin{figure}[t]
  \centering
  \includegraphics[width=\linewidth]{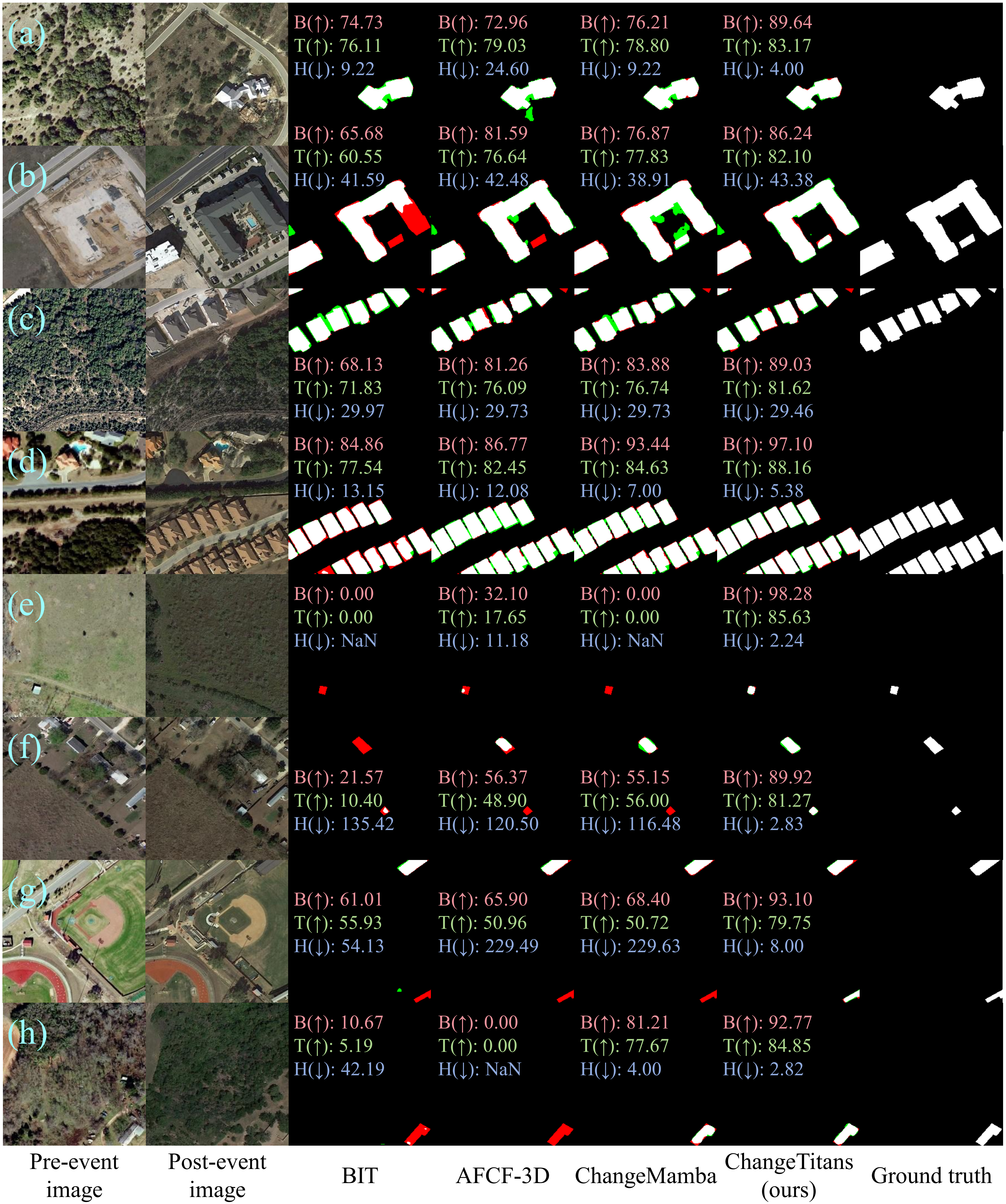}
  \caption{Qualitative results on the LEVIR-CD \cite{chen2020spatial} dataset. 
Examples are grouped into four challenging scenarios: 
(a) \& (b) irregularly shaped objects, 
(c) \& (d) densely packed regions, 
(e) \& (f) small-scale changes, and 
(g) \& (h) changes near image boundaries. 
Predicted outputs are color-coded as follows: 
white for true positives (TP), 
black for true negatives (TN), 
\textcolor[RGB]{10,252,74}{green} for false positives (FP), 
and \textcolor[RGB]{255,11,0}{red} for false negatives (FN). 
In the figure, \textcolor[RGB]{239,148,158}{B} represents Boundary F1 score, \textcolor[RGB]{172,215,142}{T} represents Trimap-based mIoU, and \textcolor[RGB]{145,172,224}{H} represents Hausdorff distance.
}
  \label{fig:vis}
  \vspace{-0.3cm}
\end{figure}

\setlength{\fboxrule}{2pt}

\begin{figure}[t]
  \centering
  \includegraphics[width=1.0\linewidth]{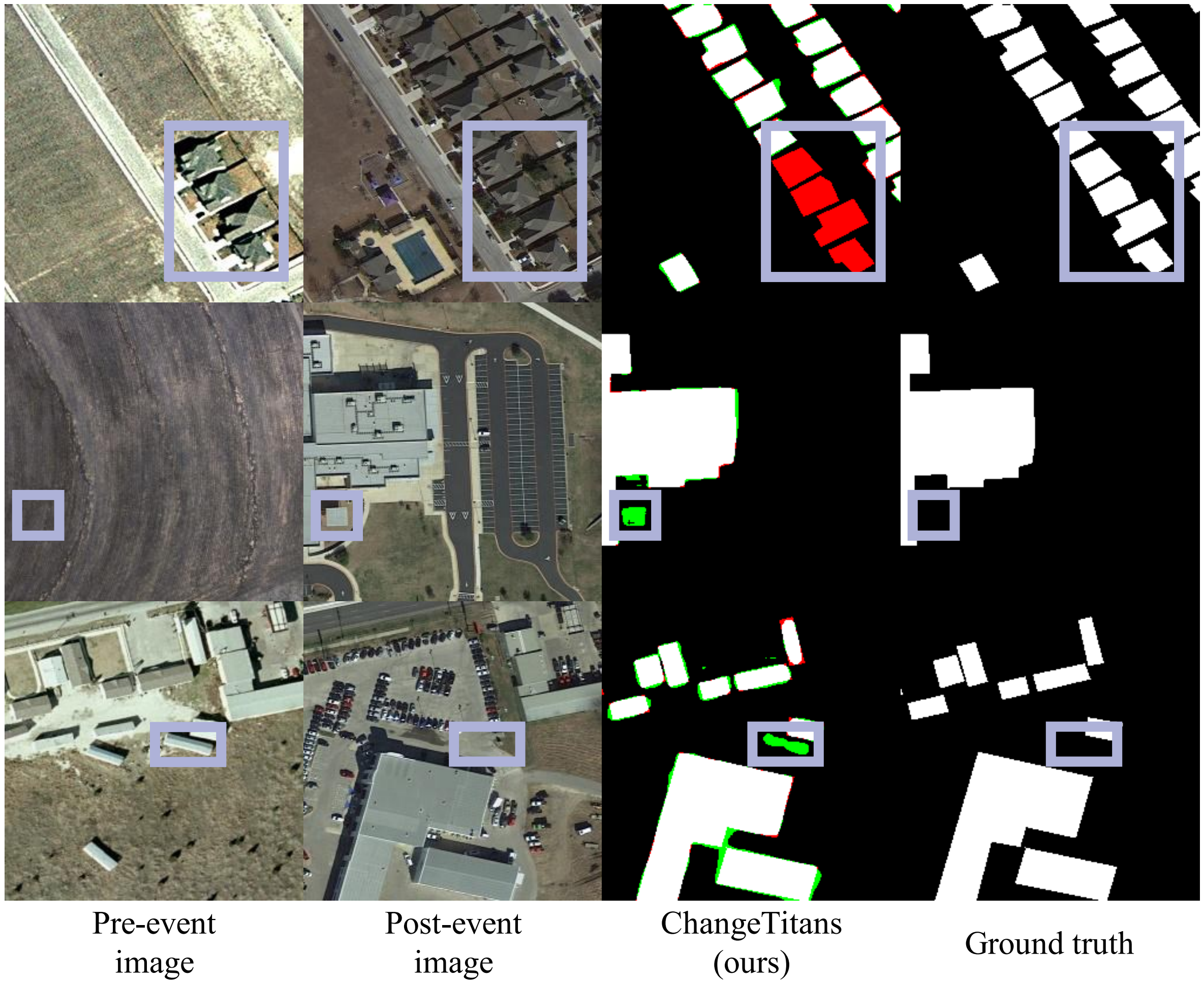}
  \caption{Qualitative comparison of change detection results on the LEVIR-CD \cite{chen2020spatial} dataset under label noise. Here, label noise refers to incomplete or incorrect annotations in the training set, as highlighted in the \fcolorbox{customPurple}{white}{\textcolor{black}{purple}} boxed regions.}
  \label{fig:noise}
  \vspace{-0.2cm}
\end{figure}

\subsection{Vision Backbone Pre-train}

During the pre-training phase, we follow the experimental setup established in previous works \cite{liu2022convnet,woo2023convnext} and pre-train our vision backbone, VTitans, on the ImageNet-1K dataset \cite{deng2009imagenet}. As reported in Table~\ref{tab:conf}, the model is trained using an input resolution of $224 \times 224$, with each image first tokenized into 16×16 non-overlapping patches, resulting in a sequence length of 196 tokens per image.
VTitans is composed of 12 stacked Titans blocks (\textit{i.e.}, $L=12$), with a hidden embedding dimension of 192. A neural memory module is inserted every 3 blocks, corresponding to layers 3, 6, 9, and 12. The chunk size for segmented attention is set to 64, meaning attention is computed independently in fixed-length sequences of 64 tokens to reduce computational cost.
The model is trained using the Adam optimizer \cite{kingma2014adam}, with a base learning rate of 4e-3, cosine annealing learning rate decay, and a batch size of 64 per GPU across 4 NVIDIA RTX 3090 GPUs. A linear warm-up phase is applied for the first 20 epochs to stabilize convergence, after which the learning rate follows the cosine schedule. Gradient clipping with a maximum norm of 1.0 is applied to prevent exploding gradients due to the memory update mechanism. No Exponential Moving Average (EMA) is used during pre-training.

As reported in Table~\ref{tab:pretrain}, VTitans achieves a Top-1 accuracy of 75.7\% on ImageNet-1K, demonstrating competitive performance relative to similar scale models.
Compared to ResNet-18, a commonly adopted in remote sensing change detection models such as STANet-PAM \cite{chen2020spatial}, ChangeStar \cite{zheng2021change}, and Changer \cite{fang2023changer}, VTitans improves Top-1 accuracy by 5.8\% with an equivalent parameter budget. These results highlight the effectiveness of the integrated neural memory mechanism. Unlike CNN-based models, VTitans significantly enhances global contextual modeling through memory-based representations. In contrast to Transformer-based methods, VTitans combines segmented attention for local feature extraction with memory-based global reasoning, delivering strong performance while maintaining architectural efficiency.

\subsection{Experiments on LEVIR-CD and WHU-CD}

\textbf{Quantitative Results}.
We comprehensively evaluate ChangeTitans on two popular benchmarks, LEVIR-CD and WHU-CD, to highlight its effectiveness in remote sensing change detection. As presented in Table \ref{tab:levir}, our method achieves strong results on both datasets while maintaining a relatively modest model size (27.15M Params) and computational cost (30.39G FLOPs). On LEVIR-CD, ChangeTitans attains an IoU of 84.36\% and an F1-score of 91.52\%, surpassing leading methods such as ChangeMamba (84.27\% IoU, 91.37\% F1) and ChangeCLIP (83.99\% IoU, 91.30\% F1). Notably, our model achieves a recall of 90.56\%, a critical metric in change detection where missed events can severely impact downstream applications. On WHU-CD, ChangeTitans reaches an F1-score of 94.10\% and an IoU of 88.56\%, matching or exceeding recent top performers. While ChangeCLIP reports a marginally higher F1 (94.78\%), it relies on a more complex two-stage pipeline, whereas ChangeTitans delivers comparable or better performance under a single, unified framework. These results firmly establish the robustness and practicality of our approach in detecting diverse changes with high accuracy and efficiency.

\begin{table}[t]
\caption{Comparison results on the LEVIR-CD+ \cite{shen2021s2looking} dataset. All metrics are reported as percentages (\%), with the highest value highlighted in \textbf{bold}.}
\label{tab:levirp}
\setlength{\tabcolsep}{3pt}
\resizebox{\columnwidth}{!}{%
\begin{tabular}{lccccc}
\specialrule{1pt}{0pt}{0pt}
\rowcolor[HTML]{D6D8EB} 
\multicolumn{1}{c}{\cellcolor[HTML]{D6D8EB}\textbf{Method}} &
  \textbf{Venue} &
  \textbf{\begin{tabular}[c]{@{}c@{}}Params\\ (M)\end{tabular}} &
  \textbf{\begin{tabular}[c]{@{}c@{}}FLOPs\\ (G)\end{tabular}} &
  \textbf{IoU} &
  \textbf{F1} \\ \hline
\rowcolor[HTML]{EFEFEF} 
FC-Siam-Diff \cite{daudt2018fully}        & ICIP 2018  & 4.38  & 1.35   & 57.44 & 72.97 \\
\rowcolor[HTML]{EFEFEF} 
FC-Siam-Conc \cite{daudt2018fully}        & ICIP 2018  & 4.99  & 1.55   & 58.07 & 73.48 \\
\rowcolor[HTML]{EFEFEF} 
STANet \cite{chen2020spatial}             & RS 2020    & 16.93 & 6.58   & 65.66 & 79.31 \\
\rowcolor[HTML]{EFEFEF} 
SNUNet \cite{fang2021snunet}              & GRSL 2021  & 12.03 & 27.44  & 67.11 & 80.32 \\
\rowcolor[HTML]{EFEFEF} 
BIT \cite{chen2021remote}                 & TGRS 2021  & 3.55  & 4.35   & 70.64 & 82.80 \\
\rowcolor[HTML]{EFEFEF} 
TransUNetCD \cite{li2022transunetcd}      & TGRS 2022  & 28.37 & 244.54 & 71.86 & 83.63 \\
\rowcolor[HTML]{EFEFEF} 
SwinSUNet \cite{zhang2022swinsunet}       & TGRS 2022  & 39.28 & 43.50  & 74.82 & 85.60 \\
\rowcolor[HTML]{EFEFEF} 
CTDFormer \cite{zhang2023relation}        & TGRS 2023  & 3.85  & 303.77 & 67.09 & 80.30 \\
\rowcolor[HTML]{EFEFEF} 
BiFA \cite{zhang2024bifa}                 & TGRS 2024  & 5.58  & 53.00  & 72.35 & 83.96 \\
\rowcolor[HTML]{EFEFEF} 
ChangeCLIP \cite{dong2024changeclip}      & ISPRS 2024 & -     & -      & 75.63 & 86.12 \\
\rowcolor[HTML]{EFEFEF} 
ConvFormer-CD/48 \cite{yang2025convformer} & TGRS 2025 & 37.72 & 5.14   & 74.37 & 85.30 \\
\rowcolor[HTML]{EFEFEF} 
SPMNet \cite{wang2025spmnet}              & TGRS 2025  & 17.33 & 6.94   & 71.01 & 83.09 \\ \hline
\rowcolor[HTML]{D6D8EB} 
\textbf{ChangeTitans (ours)}              & -          & 27.15 & 30.39  & \textbf{75.84} & \textbf{86.37} \\ \specialrule{1pt}{0pt}{0pt}
\end{tabular}%
}
\end{table}

\begin{table}[t]
\caption{Comparison results on the SYSU-CD \cite{shi2021deeply} dataset. All metrics are reported as percentages (\%), with the highest value highlighted in \textbf{bold}.}
\label{tab:sysu}
\setlength{\tabcolsep}{3pt}
\resizebox{\columnwidth}{!}{%
\begin{tabular}{lccccc}
\specialrule{1pt}{0pt}{0pt}
\rowcolor[HTML]{D6D8EB} 
\multicolumn{1}{c}{\cellcolor[HTML]{D6D8EB}\textbf{Method}} &
  \textbf{Venue} &
  \textbf{\begin{tabular}[c]{@{}c@{}}Params\\ (M)\end{tabular}} &
  \textbf{\begin{tabular}[c]{@{}c@{}}FLOPs\\ (G)\end{tabular}} &
  \textbf{IoU} &
  \textbf{F1} \\ \hline
\rowcolor[HTML]{EFEFEF} 
FC-Siam-Diff \cite{daudt2018fully}       & ICIP 2018 & 4.38  & 1.35   & 56.96 & 72.57 \\
\rowcolor[HTML]{EFEFEF} 
FC-Siam-Conc \cite{daudt2018fully}       & ICIP 2018 & 4.99  & 1.55   & 61.75 & 76.35 \\
\rowcolor[HTML]{EFEFEF} 
STANet \cite{chen2020spatial}            & RS 2020   & 16.93 & 6.58   & 63.09 & 77.37 \\
\rowcolor[HTML]{EFEFEF} 
SNUNet \cite{fang2021snunet}             & GRSL 2021 & 12.03 & 27.44  & 66.35 & 79.77 \\
\rowcolor[HTML]{EFEFEF} 
BIT \cite{chen2021remote}                & TGRS 2021 & 3.55  & 4.35   & 61.39 & 76.08 \\
\rowcolor[HTML]{EFEFEF} 
ICIFNet \cite{feng2022icif}              & TGRS 2022 & 23.82 & 25.36  & 68.12 & 80.74 \\
\rowcolor[HTML]{EFEFEF} 
TransUNetCD \cite{li2022transunetcd}     & TGRS 2022 & 28.37 & 244.54 & 66.79 & 80.09 \\
\rowcolor[HTML]{EFEFEF} 
SwinSUNet \cite{zhang2022swinsunet}      & TGRS 2022 & 39.28 & 43.50  & 68.89 & 81.58 \\
\rowcolor[HTML]{EFEFEF} 
CTDFormer \cite{zhang2023relation}       & TGRS 2023 & 3.85  & 303.77 & 64.04 & 78.08 \\
\rowcolor[HTML]{EFEFEF} 
WNet \cite{tang2023wnet}                 & TGRS 2023 & 43.07 & 19.20  & 67.55 & 80.64 \\
\rowcolor[HTML]{EFEFEF} 
A2Net \cite{li2023lightweight}           & TGRS 2023 & 3.78  & 6.02   & 70.86 & 81.06 \\
\rowcolor[HTML]{EFEFEF} 
SEIFNet \cite{huang2024spatiotemporal}   & TGRS 2024 & 8.37  & 27.91  & 69.96 & 82.32 \\
\rowcolor[HTML]{EFEFEF} 
MutSimNet \cite{liu2024mutsimnet}        & TGRS 2024 & 45.44 & 58.10  & 70.27 & 82.34 \\
\rowcolor[HTML]{EFEFEF} 
STENet \cite{pan2024stenet}              & TGRS 2024 & 10.95 & 14.73  & 69.92 & 82.29 \\ 
\rowcolor[HTML]{EFEFEF} 
ChangeMamba \cite{chen2024changemamba}   & TGRS 2024 & 84.70 & 179.32 & 71.10 & 83.11 \\ 
\rowcolor[HTML]{EFEFEF} 
AMDANet \cite{su2025amdanet}             & TGRS 2025 & 25.83 & 77.23  & 67.82 & 80.83 \\
\rowcolor[HTML]{EFEFEF} 
ChangeDA \cite{meng2025changeda}         & TGRS 2025 & 30.25 & 23.49  & 70.56 & 82.74 \\
\hline
\rowcolor[HTML]{D6D8EB} 
\textbf{ChangeTitans (ours)}             & -         & 27.15 & 30.39  & \textbf{71.18} & \textbf{83.24} \\ \specialrule{1pt}{0pt}{0pt}
\end{tabular}%
}
\vspace{-0.1cm}
\end{table}

\textbf{Qualitative Results}.
Fig.~\ref{fig:vis} presents qualitative comparisons on the LEVIR-CD dataset, organized into four challenging scenarios: (a) \& (b) irregularly shaped objects, (c) \& (d) densely packed regions, (e) \& (f) small-scale changes, and (g) \& (h) changes near image boundaries. In all cases, ChangeTitans delivers clearer boundaries, more complete detections, and fewer false alarms than competing methods. Beyond standard metrics such as IoU and F1-score, we further assess boundary preservation using Boundary F1 score, Trimap-based mIoU, and Hausdorff distance. Consistent improvements on these indicators validate that ChangeTitans not only enhances semantic consistency but also produces spatially precise change maps, particularly along complex or ambiguous object boundaries.
In real-world datasets, partial or incomplete labeling in the training set often introduces noise that can mislead model optimization \cite{sheng2024enhancing}. As shown in Fig.~\ref{fig:noise}, the boxed regions highlight such mislabeled or unannotated areas. Some baseline methods overfit to these errors, either missing true changes or being suppressed by noisy supervision. In contrast, ChangeTitans leverages neural memory to capture global context, which allows it to accurately recover the full extent of changes while maintaining robustness against imperfect annotations.

\begin{figure}[t]
  \centering
  \includegraphics[width=\linewidth]{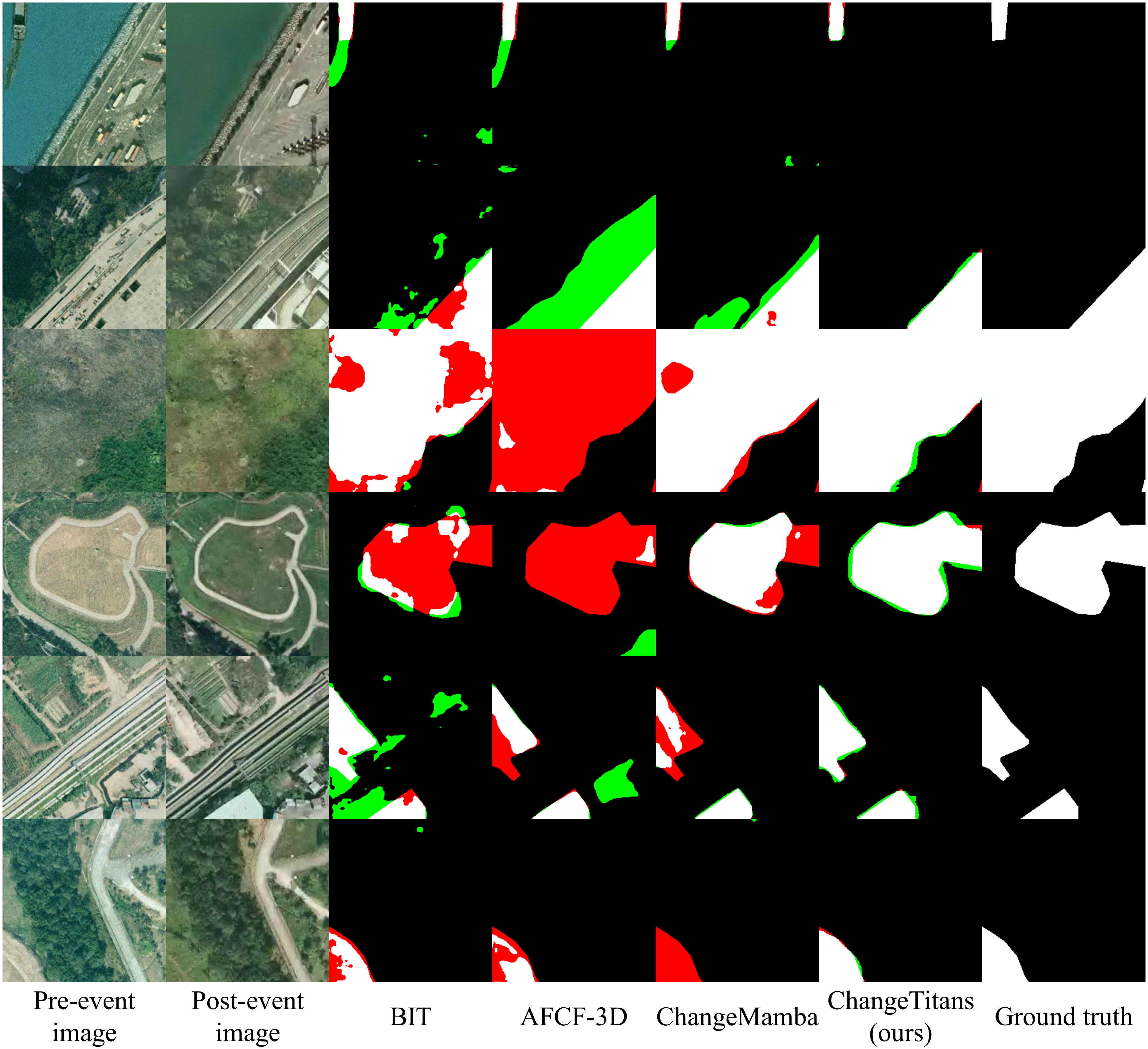}
  \caption{Qualitative results on the SYSU-CD \cite{shi2021deeply} datasets. Predicted outputs are color-coded as follows: white for true positives (TP), black for true negatives (TN), \textcolor[RGB]{10,252,74}{green} for false positives (\textcolor[RGB]{10,252,74}{FP}), and \textcolor[RGB]{255,11,0}{red} for false negatives (\textcolor[RGB]{255,11,0}{FN}).}
  \label{fig:sysu}
  \vspace{-0.1cm}
\end{figure}

\subsection{Experiments on LEVIR-CD+ and SYSU-CD}

\textbf{Quantitative Results}.
We evaluate ChangeTitans against a range of representative baselines on the LEVIR-CD+ and SYSU-CD datasets, using IoU and F1-score as evaluation metrics. As shown in Tables~\ref{tab:levirp} and \ref{tab:sysu}, our method consistently achieves top-tier performance. Specifically, ChangeTitans attains an IoU of 75.84\% and an F1-score of 86.37\% on LEVIR-CD+, and 71.18\% IoU and 83.24\% F1-score on SYSU-CD, outperforming all compared methods.
These improvements arise from the proposed model’s capacity to model long-range spatial dependencies through the proposed neural memory, which maintains and updates contextual representations across the image. Combined with segmented attention for local refinement, ChangeTitans achieves high recall and precise boundary localization, enabling robust and accurate change detection across diverse scenarios.

\textbf{Qualitative results}.  
To further assess the generalization ability of the proposed method beyond building-related changes, we present qualitative comparisons on the SYSU-CD dataset in Fig.~\ref{fig:sysu}. This dataset contains diverse types of changes, such as vegetation growth and land-cover transitions, which differ significantly from man-made structures. As shown in Fig.~\ref{fig:sysu}, our method produces clearer boundaries and reduces spurious predictions compared with baseline approaches, particularly in regions with irregular object shapes and heterogeneous backgrounds. These results complement the quantitative findings in Table~\ref{tab:sysu}, confirming that the proposed framework is also effective in non-building change scenarios.

\begin{table}[t]
\caption{Comparison results on the SAR-CD \cite{alatalo2023improved} dataset. All metrics are reported as percentages (\%), with the highest value highlighted in \textbf{bold}.}
\label{tab:sar}
\setlength{\tabcolsep}{3pt}
\resizebox{\columnwidth}{!}{%
\begin{tabular}{lccccc}
\specialrule{1pt}{0pt}{0pt}
\rowcolor[HTML]{D6D8EB} 
\textbf{Method}             & \textbf{Venue}       & \begin{tabular}[c]{@{}c@{}}\textbf{Params}\\ \textbf{(M)}\end{tabular} & \begin{tabular}[c]{@{}c@{}}\textbf{FLOPs}\\ \textbf{(G)}\end{tabular} & \textbf{IoU}   & \textbf{F1}    \\ \hline
\rowcolor[HTML]{EFEFEF} 
FC-SiamDiff \cite{daudt2018fully}       & ICIP 2018   & 4.38  & 1.35   & 86.07 & 92.52 \\
\rowcolor[HTML]{EFEFEF} 
FC-SiamConc \cite{daudt2018fully}       & ICIP 2018   & 4.99  & 1.55   & 93.39 & 96.58 \\
\rowcolor[HTML]{EFEFEF} 
BIT \cite{chen2021remote}               & TGRS 2021   & 3.55  & 4.35   & 93.48 & 96.63 \\
\rowcolor[HTML]{EFEFEF} 
MSCANet \cite{liu2022cnn}               & JSTARS 2022 & 16.59 & 14.68  & 94.43 & 97.14 \\
\rowcolor[HTML]{EFEFEF} 
FCS \cite{chen2022fccdn}                & ISPRS 2022  & 3.03  & 9.59   & 94.77 & 97.31 \\
\rowcolor[HTML]{EFEFEF} 
DED \cite{chen2022fccdn}                & ISPRS 2022  & 3.10  & 9.90   & 94.94 & 97.40 \\
\rowcolor[HTML]{EFEFEF} 
FCCDN \cite{chen2022fccdn}              & ISPRS 2022  & 6.31  & 12.52  & 95.15 & 97.51 \\
\rowcolor[HTML]{EFEFEF} 
AFCF-3D \cite{ye2023adjacent}           & TGRS 2023   & 17.54 & 31.72  & 93.72 & 96.76 \\
\rowcolor[HTML]{EFEFEF} 
ChangeMamba \cite{chen2024changemamba}  & TGRS 2024   & 84.70 & 179.32 & 90.40 & 94.96 \\ \hline
\rowcolor[HTML]{D6D8EB} 
\textbf{ChangeTitans(ours)}             & -           & 27.15 & 30.39  & \textbf{95.63} & \textbf{97.76} \\ \specialrule{1pt}{0pt}{0pt}
\end{tabular}
}
\vspace{-0.1cm}
\end{table}

\begin{figure}[t]
  \centering
  \includegraphics[width=\linewidth]{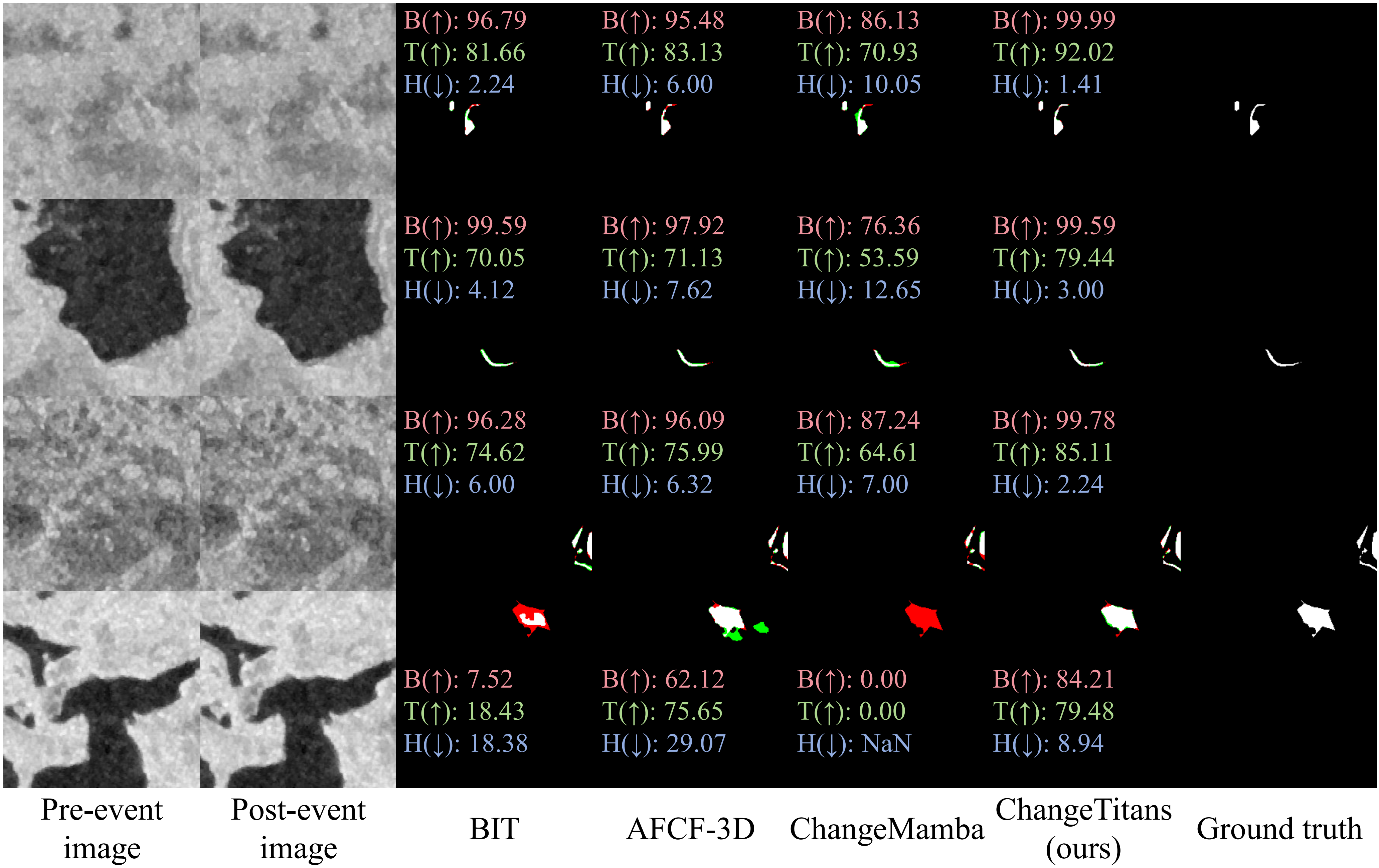}
  \caption{Qualitative results on the SAR-CD \cite{alatalo2023improved} datasets. Predicted outputs are color-coded as follows: white for true positives (TP), black for true negatives (TN), \textcolor[RGB]{10,252,74}{green} for false positives (\textcolor[RGB]{10,252,74}{FP}), and \textcolor[RGB]{255,11,0}{red} for false negatives (\textcolor[RGB]{255,11,0}{FN}). In the figure, \textcolor[RGB]{239,148,158}{B} represents Boundary F1 score, \textcolor[RGB]{172,215,142}{T} represents Trimap-based mIoU, and \textcolor[RGB]{145,172,224}{H} represents Hausdorff distance.}
  \label{fig:sar}
  \vspace{-0.2cm}
\end{figure}

\subsection{Experiments on SAR-CD}

\textbf{Quantitative Results}. We further evaluate the proposed framework on the SAR-CD dataset, which poses additional challenges due to strong speckle noise and cross-sensor variations. As reported in Table~\ref{tab:sar}, ChangeTitans achieves the best performance among all compared methods, reaching an IoU of \textbf{95.63}\% and an F1-score of \textbf{97.76}\%. These improvements are consistent across both metrics and surpass recent transformer- and convolution-based approaches, such as FCCDN~\cite{chen2022fccdn} and MSCANet~\cite{liu2022cnn}. Notably, the proposed method attains these gains while maintaining competitive parameter and computational complexity, highlighting its robustness and efficiency in handling challenging SAR change detection scenarios.

\textbf{Qualitative results}.  
To complement the quantitative analysis, Fig.~\ref{fig:sar} presents qualitative comparisons on the SAR-CD dataset. Due to the inherent speckle noise and radiometric variations in SAR imagery, many existing methods tend to produce scattered false alarms or blurred boundaries. In contrast, our method yields cleaner change maps with more precise delineation of object contours, even in cluttered or low-contrast regions. 
These visual results further demonstrate the robustness of the proposed framework in handling the unique challenges of SAR-based change detection.

\begin{table}[t]
\caption{Ablation studies on the LEVIR-CD dataset. Rows highlighted in \colorbox{customPurple}{purple} indicate the default configuration. The highest value is highlighted in \textbf{bold}.}
\label{tab:ablation}
\setlength{\tabcolsep}{5pt}
\resizebox{\columnwidth}{!}{%
\begin{tabular}{
>{\columncolor[HTML]{ECF4FF}}c
>{\columncolor[HTML]{EFEFEF}}c 
>{\columncolor[HTML]{EFEFEF}}c 
>{\columncolor[HTML]{EFEFEF}}c 
>{\columncolor[HTML]{EFEFEF}}c 
>{\columncolor[HTML]{EFEFEF}}c }
\specialrule{1pt}{0pt}{0pt}
\textbf{Experiment}                                                                      & \cellcolor[HTML]{D6D8EB}\textbf{Method} & \cellcolor[HTML]{D6D8EB}\textbf{IoU}   
                                                                                         & \cellcolor[HTML]{D6D8EB}\textbf{F1}     & \cellcolor[HTML]{D6D8EB}\textbf{P}     
                                                                                         & \cellcolor[HTML]{D6D8EB}\textbf{R}     \\ \hline
                                                                                         & \textit{w}/\textit{o} & 81.88           & 90.04               & 90.83            & 89.26\\
\multirow{-2}{*}{\textbf{\begin{tabular}[c]{@{}c@{}}Neural\\ Memory\end{tabular}}}       & \cellcolor[HTML]{D6D8EB}\textit{w}/     & \cellcolor[HTML]{D6D8EB}\textbf{84.36} 
                                                                                         & \cellcolor[HTML]{D6D8EB}\textbf{91.52}  & \cellcolor[HTML]{D6D8EB}\textbf{92.49} 
                                                                                         & \cellcolor[HTML]{D6D8EB}\textbf{90.56} \\ \hline
                                                                                         & 128            & 83.63                  & 91.08               & 91.21            & 90.96\\
                                                                                         & \cellcolor[HTML]{D6D8EB}64              & \cellcolor[HTML]{D6D8EB}84.36          
                                                                                         & \cellcolor[HTML]{D6D8EB}91.52           & \cellcolor[HTML]{D6D8EB}92.49          
                                                                                         & \cellcolor[HTML]{D6D8EB}\textbf{90.56}          \\
\multirow{-3}{*}{\textbf{\begin{tabular}[c]{@{}c@{}}Chunk\\ Size\end{tabular}}}          & 32             & \textbf{84.45}         & \textbf{91.57}      & \textbf{92.77}   & 90.41\\ \hline
                                                                                         & \textit{w}/\textit{o} & 80.45           & 89.16               & 90.24            & 88.12\\
                                                                                         & Semi           & 83.30                  & 90.89               & 92.30            & 89.52\\
\multirow{-3}{*}{\textbf{Adapter}}                                                       & \cellcolor[HTML]{D6D8EB}\textit{w}/     & \cellcolor[HTML]{D6D8EB}\textbf{84.36} 
                                                                                         & \cellcolor[HTML]{D6D8EB}\textbf{91.52}  & \cellcolor[HTML]{D6D8EB}\textbf{92.49}          
                                                                                         & \cellcolor[HTML]{D6D8EB}\textbf{90.56}  \\ \hline
                                                                                         & Early Fusion   & 80.28                  & 89.06               & 90.39            & 87.78\\
                                                                                         & Siam-Diff \cite{daudt2018fully} & 81.27 & 89.66               & 90.25            & 89.09\\
                                                                                         & Siam-Conc \cite{daudt2018fully} & 81.56 & 89.85               & 91.06            & 88.67\\
                                                                                         & FHD \cite{pei2022feature}       & 83.20 & 90.83               & 90.74            & 90.92\\
                                                                                         & STRM \cite{chen2024changemamba} & 81.81 & 90.00               & 91.49            & 88.55\\
                                                                                         & TS-CBAM (Conv) & 83.85                  & 91.21               & 90.98    & \textbf{91.44}\\
                                                                                         & TS-CBAM (Diff) & 84.12                  & 91.38               & 91.55            & 91.20\\
\multirow{-8}{*}{\textbf{\begin{tabular}[c]{@{}c@{}}Fusion\\ Module\end{tabular}}}       & \cellcolor[HTML]{D6D8EB}TS-CBAM (Sum)   & \cellcolor[HTML]{D6D8EB}\textbf{84.36}          
                                                                                         & \cellcolor[HTML]{D6D8EB}\textbf{91.52}  & \cellcolor[HTML]{D6D8EB}\textbf{92.49}          
                                                                                         & \cellcolor[HTML]{D6D8EB}90.56           
                                                                                         \\ \hline
                                                                                         & $\lambda=0.5$  &  83.76         & 91.16       & 91.96        & 90.39 \\
                                                                                         & \cellcolor[HTML]{D6D8EB}$\lambda=1.0$          
                                                                                         & \cellcolor[HTML]{D6D8EB}\textbf{84.36}          
                                                                                         & \cellcolor[HTML]{D6D8EB}\textbf{91.52} 
                                                                                         & \cellcolor[HTML]{D6D8EB}\textbf{92.49}          
                                                                                         & \cellcolor[HTML]{D6D8EB}90.56  
                                                                                         \\
\multirow{-3}{*}{\textbf{\begin{tabular}[c]{@{}c@{}}Balance\\ Factor \end{tabular}}}     & $\lambda=2.0$  &  84.20         & 91.42       & 91.80        & \textbf{91.05} \\ \hline
                                                                                         & Bilinear       & 83.84                  & 91.21               & 92.00            & 90.42 \\
\multirow{-2}{*}{\textbf{\begin{tabular}[c]{@{}c@{}}Upsampling\\ Strategy\end{tabular}}} & \cellcolor[HTML]{D6D8EB}Convex          & \cellcolor[HTML]{D6D8EB}\textbf{84.36}          
                                                                                         & \cellcolor[HTML]{D6D8EB}\textbf{91.52}  & \cellcolor[HTML]{D6D8EB}\textbf{92.49}          
                                                                                         & \cellcolor[HTML]{D6D8EB}\textbf{90.56}  \\ \specialrule{1pt}{0pt}{0pt}
\end{tabular}%
}
\end{table}

\subsection{Ablation Study}

We conduct an extensive ablation study on the LEVIR-CD dataset to evaluate the individual contributions of key components within ChangeTitans, including neural memory, memory chunk size, the adapter module, fusion strategies, balance factor, and upsampling strategy. Table~\ref{tab:ablation} summarizes the results, which consistently indicate that each proposed element enhances change detection accuracy.

\noindent\textbf{Neural memory}.
We first evaluate the impact of neural memory. As shown in Table~\ref{tab:ablation}, incorporating neural memory significantly improves across all evaluation metrics. Specifically, without memory, the model’s IoU and F1-score drop from 84.36\% to 81.88\% and from 91.52\% to 90.04\%, respectively. This underscores the importance of memory-based global context modeling. By retaining and updating contextual cues, the neural memory enables ChangeTitans to distinguish subtle or overlapping changes more accurately.

\begin{figure}[t]
  \centering
  \includegraphics[width=\linewidth]{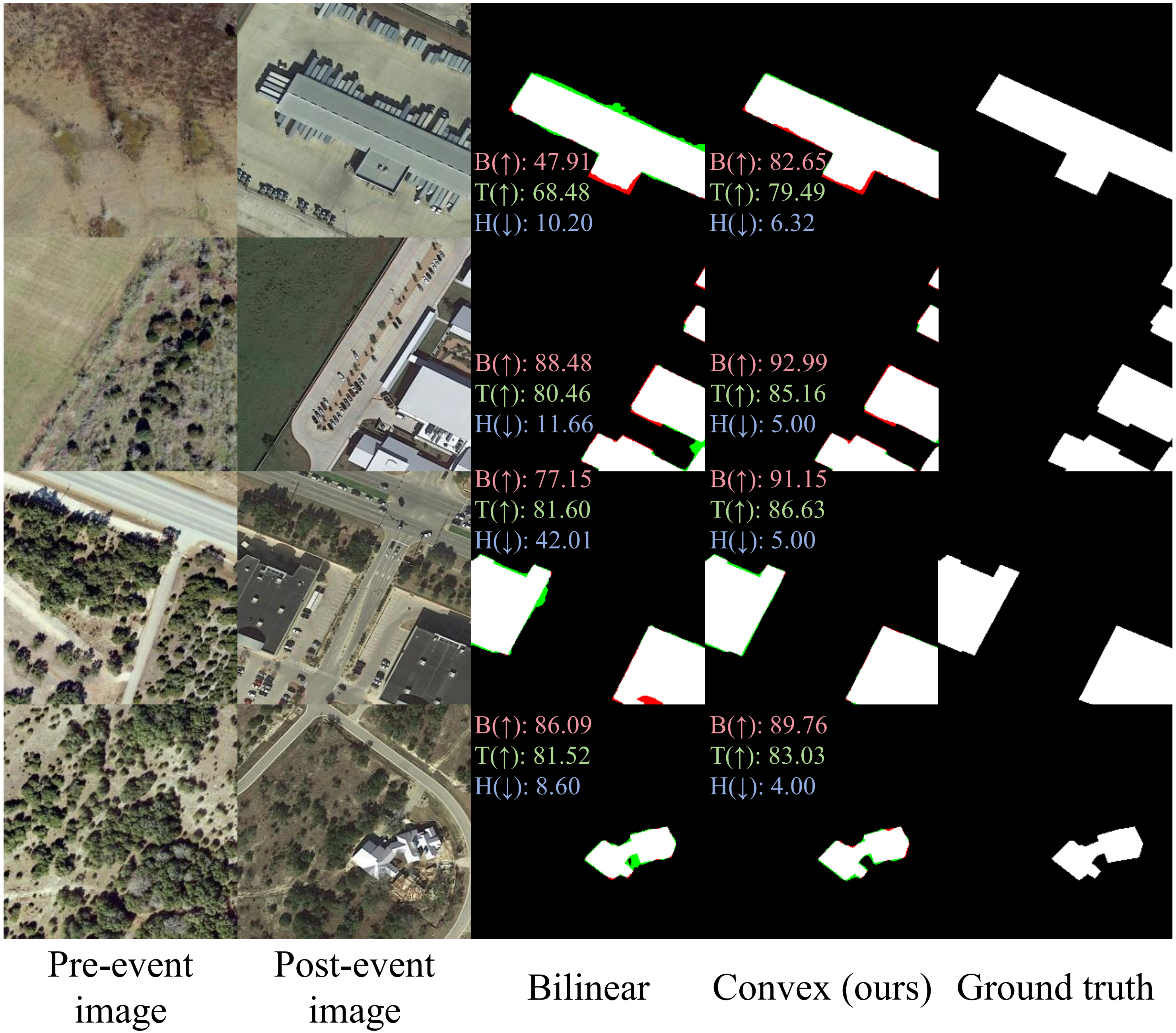}
  \caption{Qualitative comparison of bilinear interpolation and our convex upsampling on the LEVIR-CD dataset. Predicted outputs are color-coded as follows: white for true positives (TP), black for true negatives (TN), \textcolor[RGB]{10,252,74}{green} for false positives (\textcolor[RGB]{10,252,74}{FP}), and \textcolor[RGB]{255,11,0}{red} for false negatives (\textcolor[RGB]{255,11,0}{FN}). In the figure, \textcolor[RGB]{239,148,158}{B} represents Boundary F1 score, \textcolor[RGB]{172,215,142}{T} represents Trimap-based mIoU, and \textcolor[RGB]{145,172,224}{H} represents Hausdorff distance.}
  \label{fig:up}
  \vspace{-0.2cm}
\end{figure}

\noindent\textbf{Chunk size}.
We then vary the chunk size in the neural memory. As shown in Table~\ref{tab:ablation}, a chunk size of 64 yields the best trade-off between performance and memory usage, achieving 84.36\% IoU and 91.52\% F1-score. A larger chunk size (128) reduces performance, while finer segmentation offers only marginal gains at the cost of higher memory consumption. These observations are consistent with Titans \cite{behrouz2024titans}, confirming that chunk size is a key factor in balancing efficiency and accuracy.

\noindent\textbf{VTitans-Adapter}.
To assess hierarchical feature encoding, we compare three Adapter configurations in Table \ref{tab:ablation}: without Adapter, Semi-Adapter\footnote{Only shallow features are refined into a higher-resolution hierarchical form, and deeper features remain at their original resolution \(\bigl( \textit{i.e.}, f_{i4}^\prime \in \mathbb{R}^{(H/p)\times (W/p)\times C}\bigr)\).}, and the full Adapter. Without the Adapter, the model reaches 80.45\% IoU and 89.16\% F1-score, indicating that flat representations are suboptimal for change detection. Semi-Adapter significantly boosts performance to 83.30\% IoU and 90.89\% F1-score, while the full Adapter achieves 84.36\% IoU and 91.52\% F1-score alongside a recall of 90.56\%. This demonstrates that transforming non-hierarchical features into a well-structured, multi-scale representation substantially aids fine-grained change localization.

\noindent\textbf{Fusion module}.
As reported in Table \ref{tab:ablation}, we compare various bitemporal fusion strategies, including early fusion (channel-wise concatenation of input images), Siamese-based methods such as Siam-Diff \cite{daudt2018fully} and Siam-Conc \cite{daudt2018fully}, FHD \cite{pei2022feature}, and STRM from ChangeMamba \cite{chen2024changemamba}. We then evaluate three variants of our TS-CBAM (Diff, Sum, Conv), each performing a different operation on the fused features. TS-CBAM (Sum) achieves the highest IoU of 84.36\% and F1-score of 91.52\%, indicating that summation effectively preserves key information while minimizing background noise. These results confirm the advantage of TS-CBAM over classical fusion modules, particularly in distinguishing subtle or partial changes.

\noindent\textbf{Balance factor}.
We further examine the effect of balancing the BCE and Dice losses by introducing a factor \(\lambda\), which controls the contribution of the Dice loss. When \(\lambda = 1\), equal importance is assigned to both loss terms. Empirical results show that this setting achieves the best overall performance, suggesting that jointly optimizing for semantic consistency (Dice) and per-pixel accuracy (BCE) leads to more stable training and precise change detection.

\noindent\textbf{Upsampling strategy}.  
To evaluate the contribution of the Convex Upsampling module, we conduct an ablation study by replacing it with standard bilinear interpolation in the decoder. As reported in Table~\ref{tab:ablation}, Convex Upsampling consistently improves performance across key metrics, with IoU increasing from 83.84\% to 84.36\%, and F1-score improving from 91.21\% to 91.52\%.  
In addition to quantitative gains, we provide a qualitative comparison in Fig.~\ref{fig:up}. The results illustrate that bilinear interpolation tends to produce blurred or misaligned boundaries, whereas Convex Upsampling yields sharper and more reliable delineation of change regions. This is further supported by boundary-sensitive metrics, which highlight improvements along irregular object edges and small-scale structures.
Unlike fixed interpolation schemes, Convex Upsampling adaptively aggregates local low-resolution features under a convex constraint, enabling structure-aware reconstruction and enhancing robustness in challenging remote sensing scenarios.

\section{Conclusion}

This paper presents ChangeTitans, a remote sensing change detection framework that builds on the Titans architecture with neural memory to address the challenges of capturing long-range dependencies while maintaining computational efficiency. By integrating the VTitans backbone, hierarchical feature adaptation, and the TS-CBAM fusion module, the model effectively captures both local and global contextual information while reducing spurious variations from seasonal or lighting differences. Extensive experiments on four benchmark datasets (LEVIR-CD, WHU-CD, LEVIR-CD+, and SYSU-CD) show that ChangeTitans achieves state-of-the-art performance, surpassing existing methods in accuracy and robustness. Ablation studies confirm the importance of neural memory and adaptive feature fusion in achieving these gains.

Despite these advances, change detection in remote sensing remains challenged by persistent issues such as pseudo-change suppression and fine-grained boundary localization, particularly in scenarios with significant temporal gaps or cross-sensor imagery. Future work will improve the model's adaptability to diverse sensing environments and explore lightweight variants suitable for real-time deployment scenarios.

\bibliographystyle{IEEEtran}
\bibliography{changetitans}

\vfill

\end{document}